\documentclass[preprint, 12pt]{elsarticle}

\usepackage{hyperref, lineno}
\usepackage{amsmath, amssymb}
\usepackage{amsfonts}
\usepackage{graphicx}
\usepackage{bm, bbm}
\usepackage{algorithm, caption}
\usepackage{algpseudocode}
\usepackage{afterpage}
\usepackage{physics}
\usepackage{subcaption}
\usepackage{multirow}
\usepackage{color}
\usepackage{enumitem}\setlist[enumerate]{itemsep = 0mm}

\newcommand{\R}{\rm I\!R}

\journal{Journal Name}

\begin{document}

\begin{frontmatter}
\title{Physics-Informed Generative Adversarial Networks for Stochastic Differential Equations}
\author{Liu Yang}
\author{Dongkun Zhang}
\author{George Em Karniadakis\corref{cor}} \ead{george\_karniadakis@brown.edu}
\address{Division of Applied Mathematics, Brown University, Providence, RI 02912, USA}
\cortext[cor]{Corresponding Author}

\begin{abstract}

We developed a new class of physics-informed generative adversarial networks (PI-GANs) to solve in a unified manner forward, inverse and mixed stochastic problems based on a limited number of scattered measurements. Unlike standard GANs relying only on data for training, here we encoded into the architecture of GANs the governing physical laws in the form of stochastic differential equations (SDEs) using automatic differentiation. In particular, we applied Wasserstein GANs with gradient penalty (WGAN-GP) for its enhanced stability compared to vanilla GANs. We first tested WGAN-GP in approximating Gaussian processes of different correlation lengths based on data realizations collected from simultaneous reads at sparsely placed sensors. We obtained good approximation of the generated stochastic processes to the target ones even for a mismatch between the input noise dimensionality and the effective dimensionality of the target stochastic processes. We also studied the overfitting issue for both the discriminator and generator, and we found that overfitting occurs also in the generator in addition to the discriminator as previously reported. Subsequently, we considered the solution of elliptic SDEs requiring approximations of three stochastic processes, namely the solution, the forcing, and the diffusion coefficient. Here again we assumed data realizations collected from simultaneous reads at a limited number of sensors for the multiple stochastic processes. We used three generators for the PI-GANs, two of them were feed forward deep neural networks (DNNs) while the other one was the neural network induced by the SDE. For the case where we have one group of data, we employed one feed forward DNN as the discriminator while for the case of multiple groups of data we employed multiple discriminators in PI-GANs. We solved forward, inverse, and mixed problems without changing the framework of PI-GANs, obtaining both the means and standard deviations of the stochastic solution and the diffusion coefficient in good agreement with benchmarks. Here, we have demonstrated the effectiveness of PI-GANs in solving SDEs for up to 30 dimensions, but in principle, PI-GANs could tackle very high dimensional problems given more sensor data with low-polynomial growth in computational cost.

\end{abstract}

\begin{keyword}
WGAN-GP \sep multi-player GANs \sep high dimensional problems \sep inverse problems \sep elliptic stochastic problems
\end{keyword}

\end{frontmatter}

\section{Introduction} \label{S:1}

Generative adversarial networks (GANs) have achieved remarkable success within short time for diverse tasks of generating synthetic data, such as images~\cite{berthelot2017began, karras2017progressive, ledig2017photo, CycleGAN2017}, texts~\cite{yu2017seqgan,zhang2017adversarial,fedus2018maskgan, liang2017recurrent}, and even music~\cite{yu2017seqgan,mogren2016c,yang2017midinet,guimaraes2017objective}. GANs can learn probability distributions from data, an attribute suggestive of its potential application to modeling the {\em inherent stochasticity} and {\em extrinsic uncertainty} in physical and biological systems. However, to the best of our knowledge, there is no work explicitly encoding the known physical laws into the framework of GANs so far in the spirit of physics-informed neural networks first introduced in~\cite{MaziarParisGK17_1,MaziarParisGK17_2}. 

\begin{figure}[H]
	\centering
	\includegraphics[width=0.6\textwidth]{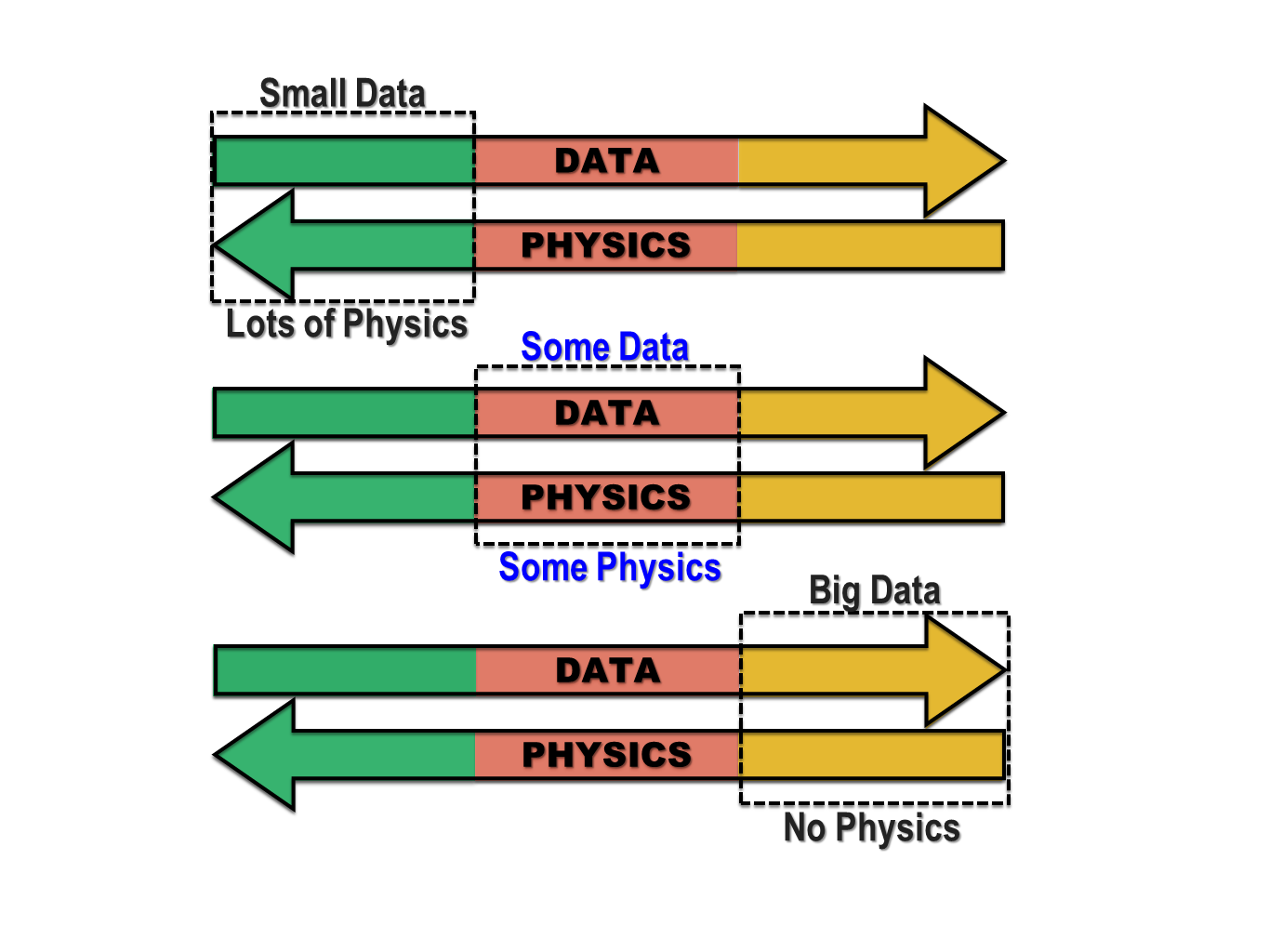}
	\caption{Schematic to illustrate three possible categories of physical problems and associated available data. We use the term ``PHYSICS'' to imply the known physics for the target problem. }
	\label{fig:Data_Physics}
s\end{figure}

The specific data-driven approach to modeling physical and biological systems depends crucially on the amount of data available as well as on the complexity of the system itself, as illustrated in Figure \ref{fig:Data_Physics}. The classical paradigm for which many different numerical methods have been developed over the last fifty years is shown on the the top of Figure \ref{fig:Data_Physics}, where we assume that the only data available are the boundary and initial conditions while the specific governing partial differential equation (PDE) and associate parameters are precisely known. On the other extreme (lower plot), we may have a lot of data, e.g. in the form of time series, but we may not know the governing physical law, e.g. the underlying PDE, at the continuum level; many problems in social dynamics fall under this category although work so far has focused on recovering known PDEs from data only, e.g. see~\cite{schmidt2009distilling,brunton2016discovering,MaziarGK18JCP}. Perhaps the most interesting category is sketched in the middle plot, where we assume that we know the physics partially, e.g. in an advection-diffusion-reaction system the reaction terms may be unknown, but we have several scattered measurements in addition to the boundary and initial conditions that we can use to infer the missing functional terms and other parameters in the PDE and simultaneously recover the solution. It is clear that this middle category is the most general case, and in fact it is representative of the other two categories, if the measurements are too few or too many. This is the {\it mixed} case that we address in this paper but with the significantly more complex scenario, where the solution is a stochastic process due to stochastic excitation or an uncertain material property, e.g. permeability or diffusivity in a porous medium. Hence, we employ stochastic differential equations (SDEs) to represent these stochastic solutions and other stochastic fields.

Taking inspiration from the work on physics-informed neural networks for deterministic PDEs~\cite{MaziarParisGK17_1,MaziarParisGK17_2}, here we employ GANs for stochastic problems in the second category of Figure \ref{fig:Data_Physics}. We wish to encode the known physics, more specifically, the {\em form of the stochastic differential equation (SDE)}, into the architecture of GANs while at the same time exploit the feature of GANs to model and learn the unknown stochastic terms in the equations from data. This approach represents a seamless integration of models and data both for inference and system identification but also for the aforementioned mixed case, where we have insufficient data for both, and hence we wish to infer both the system and the state.
%
This is an emerging new paradigm in machine learning research addressing engineering applications, where typically the physics is very complex and 
we only have partial measurements of forcing or material properties. To this end, recent advances include using Gaussian process regression~\cite{graepel2003, sarkka2011, bilionis2016, Raissi_nonlinear, Pang2018, xiuyang_GP} and deep neural networks (DNNs)~\cite{Lagaris1997, Lagaris2000, Yinglexing17, MaziarParisGK17_1, Hadi18} to solve forward problems as well as Bayesian estimation~\cite{Stuart10} and DNNs~\cite{raissi_jcp_2017,zhu_zabaras18} for inverse problems.
However, published works are mostly dealing with deterministic systems. There have only been very few works published on data-driven methods for SDEs, e.g., \cite{Weinan-arxiv, Maziar-arxiv}, for forward problems; for the inverse problem, Zhang et.al.~\cite{zhang2018quantifying} have recently proposed a DNN based method that learns the modal functions of the quantity of interest, which could also be the unknown system parameters. While robust, this method suffers from the `` curse of dimensionality'' in that the number of polynomial chaos expansion terms grows exponentially as the effective dimension increases, leading to prohibitive computational costs for modeling stochastic systems in high dimensions.
%
In this paper, we propose physics-informed GANs (PI-GANs) to solve SDEs. Our method is flexible in that without changing the general framework, it is capable of solving a wide range of problems, from forward problems to inverse problems and in between, i.e., mixed problems. Moreover, PI-GANs do not suffer from the “curse of dimensionality”. As a result, we can possibly tackle SDEs involving stochastic processes with high effective dimensions. In addition, PI-GANs can make use of data collected from multiple groups consisting of multiple sensors with no alignment between data in the groups.

The organization of this paper is as follows. In Section \ref{S:1.5}, we set up the data-driven problems. In Section \ref{S:2}, we give a brief review of GANs and the specific version we use, namely Wasserstein GANs with gradient penalty (WGAN-GP). Our main algorithms are introduced in Section \ref{sec:Methods}, followed by a detailed study of the performance of our methods in Section \ref{sec:results}. We first consider the learning of stochastic processes from a limited number of realizations and discuss the currently under-studied issue of overfitting in WGAN-GP. Subsequently, we present solutions of stochastic PDEs for different types of available data and different dimensions. We conclude in Section \ref{S:Conclusion} with a brief summary and discussion of the current limitations of the method.

\section{Problem setup}\label{S:1.5}
To illustrate the main idea of PI-GANs we consider the following stochastic differential equation:
\begin{equation} \label{eqn:General}
\begin{aligned}
    \mathcal{N}_x[u(x;\omega);k(x;\omega )] & = f(x;\omega), \quad x\in \mathcal{D}, \quad \omega \in \Omega,\\
    B_x[u(x;\omega)] & = b(x; \omega), \quad x \in \Gamma,\\
\end{aligned}
\end{equation}
where $\mathcal{N}_x$ is a general differential operator, $\mathcal{D}$ is the $d$-dimensional physical domain in $\R^d$, $\Omega$ is the probability space, and $\omega$ is a random event. The coefficient $k(x; \omega)$ and the forcing term $f(x;\omega)$ are modeled as random processes, and thus the solution $u(x;\omega)$ will depend on both $k(x; \omega)$ and $f(x;\omega)$. $B_x$ is the boundary condition operator acting on the domain boundary $\Gamma$. As a pedagogical example, in this paper we consider the one-dimensional stochastic elliptic equation, which retains most of the main features of more complex SDEs:
\begin{equation}\label{eqn:kufequation}
\begin{aligned}
	- \frac{1}{10}\frac{d}{dx}\left[k(x;\omega)\frac{d}{dx}u(x;\omega)\right] &= f(x; \omega), \quad x \in [-1,1], \\
	u(-1) & = u(1) =0,
\end{aligned}
\end{equation}
where $k(x; \omega)$ and $f(x; \omega)$ are independent stochastic processes, and $k(x;\omega)$ is strictly positive. For simplicity, we impose homogeneous Dirichlet boundary conditions on $u(x;\omega)$.

We consider a general scenario for Equation (\ref{eqn:General}), where we have a limited number of measurements from scattered sensors for the stochastic processes.
Specifically, we place sensors at $\{x^k_i\}_{i=1}^{n_k}$, $\{x^u_i\}_{i=1}^{n_u}$, $\{x^f_i\}_{i=1}^{n_f}$ and $\{x^b_i\}_{i=1}^{n_b}$ to collect ``snapshots'' of $k(x;\omega)$, $u(x;\omega)$, $f(x;\omega)$ and $b(x;\omega)$, where $n_k$, $n_u$, $n_f$ and $n_b$ are the numbers of sensors for $k(x;\omega)$, $u(x;\omega)$, $f(x;\omega)$ and $b(x;\omega)$, respectively. Here, one ``snapshot'' represents a simultaneous read of all the sensors, and we assume that the data in one snapshot correspond to the same random event $\omega$ in the random space $\Omega$, while $\omega$ varies for different snapshots. Note that each snapshot is the
 concatenation of snapshots from $k(x;\omega)$, $u(x;\omega)$, $f(x;\omega)$ and $b(x;\omega)$, thus it is actually a vector of size $(n_k + n_u + n_f + n_b)$. Suppose we have a group of $N$ snapshots, then we denote the accessible data set as $\{T(\omega^{(j)})\}_{j=1}^{N}$ defined as
\begin{equation}\label{eqn:Data}
\begin{aligned}
\{T(\omega^{(j)})\}_{j=1}^{N} & = \{(K(\omega^{(j)}), U(\omega^{(j)}), F(\omega^{(j)}), B(\omega^{(j)}))\}_{j=1}^{N},\\
K(\omega^{(j)}) &= (k(x^k_i;\omega^{(j)}))_{i=1}^{n_k}, \\
U(\omega^{(j)}) &= (u(x^u_i;\omega^{(j)}))_{i=1}^{n_u}, \\
F(\omega^{(j)}) &= (f(x^f_i;\omega^{(j)}))_{i=1}^{n_f}, \\
B(\omega^{(j)}) &= (b(x^b_i;\omega^{(j)}))_{i=1}^{n_b}. \\
\end{aligned}
\end{equation}
The corresponding terms in Equation (\ref{eqn:Data}) are omitted if we put no sensors for that specific process.

In this paper, we assume that we always have a sufficient number of sensors for $f(x;\omega)$ in Eqn \ref{eqn:kufequation}. The type of the problems varies depending on the number of sensors on $k(x;\omega)$ and $u(x;\omega)$: as we decrease the number of sensors on $k(x;\omega)$ while increase the number of sensors on $u(x;\omega)$, the problems gradually transform from forward to mixed and finally to inverse problems.



Moreover, we could have more than one group of snapshots of measurements, so in this case, our accessible data are
\begin{equation}
    \{\{T_t(\omega^{(t,j)})\}_{j=1}^{N_t}\}_{t = 1}^M = \{\{(K_t(\omega^{(t,j)}), U_t(\omega^{(t,j)}), F_t(\omega^{(t,j)}), B_t(\omega^{(t,j)}))\}_{j=1}^{N_t}\}_{t = 1}^M,
\end{equation}
where $M$ is the number of groups, $t$ is the index for the groups, and $N_t$ is the number of snapshots in group $t$. The random event $\omega^{(t,j)}$ denotes the random instance of the $j$-th snapshot in group $t$. Note that the sensor setups in different groups could be different, and we use $\{x^{k,t}_i\}_{i=1}^{n_{k,t}}$ to denote the position setup of $n_{k,t}$ sensors for $k$ in group $t$ (similarly for other terms). More importantly, snapshots from different groups could be collected independently. For example, in addition to the existing old sensors, we may place some new sensors and collect and utilize data from both the new and the old sensors. In this setup, we can make use of both the newly collected data and the previously collected data solely from the old sensors.

\section{A brief review of GANs and WGANs} \label{S:2}
Before moving to our main algorithms, we briefly review GANs and WGANs. Consider the problem of learning a distribution $P_r$ on $\R^d$, given data sampled from $P_r$. Suppose we have a DNN $G_\theta(\cdot)$ parameterized by $\theta$ that takes the random variable $z \in \R^m$ as input and outputs a sample $G_\theta(z) \in \R^d$. The random input $z$ is sampled from a prescribed distribution $Q_z$ (e.g., uniform, Gaussian), and we denote the distribution of $G_\theta(z)$ as $P_g$. We want to approximate $P_r$ with $P_g$.

GANs deal with this problem by defining a two-player zero-sum game between the generator $G_\theta(\cdot)$ and the discriminator $D_\rho(\cdot)$, which is another DNN parameterized by $\rho$. The discriminator takes a sample $x\in \R^d$ as input and aims to tell if it is sampled from $P_r$ or $P_g$. Meanwhile, the generator tries to ``deceive'' the discriminator by mimicking the true distribution $P_r$. In vanilla GANs~\cite{goodfellow2014generative}, the two-player zero-sum game is defined as follows:
\begin{equation} \label{eqn:gameforvanilla}
\inf_{\theta}\sup_{\rho}\mathbb{E}_{x\sim P_r}[\log(D_{\rho}(x))] + \mathbb{E}_{z\sim Q_z}[\log( 1 - D_{\rho}(G_{\theta}(z)))].
\end{equation}
Correspondingly, the loss functions for the generator and discriminator are
\begin{equation} \label{eqn:lossforvanilla}
\begin{aligned}
	L_g &=  \mathbb{E}_{z\sim Q_z}[\log( 1 - D_{\rho}(G_{\theta}(z)))],\\
	L_d &= -\mathbb{E}_{x\sim P_r}[\log(D_{\rho}(x))] - \mathbb{E}_{z\sim Q_z}[\log( 1 - D_{\rho}(G_{\theta}(z)))].
\end{aligned}
\end{equation}
If the discriminator is optimal, $L_g$ measures the Jensen-Shannon (JS) divergence between $P_r$ and $P_g$ up to multiplication and addition by constants:
\begin{equation} \label{eqn:JSD}
JS(P_r, P_g) = \frac{1}{2}KL(P_r||M) + \frac{1}{2}KL(P_g||M),
\end{equation}
where $M = (P_r + P_g)/2$, and $KL(\cdot||\cdot)$ is the Kullback-Leibler divergence~\cite{goodfellow2014generative}. By training the generator and discriminator iteratively, ideally we can make $P_g$ approach $P_r$ in the sense of the JS divergence. 

However, the JS divergence does not always provide a usable gradient for the generator, especially when the two distributions concentrate on low dimensional manifolds~\cite{arjovsky2017wasserstein}. As a consequence, training vanilla GANs is quite a delicate process and could be unstable~\cite{arjovsky2017wasserstein}. To fix this problem, Wasserstein GANs with clips on weights (weight-clipped WGANs)~\cite{arjovsky2017wasserstein} and WGANs with gradient penalty (WGAN-GP)~\cite{gulrajani2017improved} were proposed. Similar to vanilla GANs, the two-play zero-sum game in WGANs is formulated as
\begin{equation} \label{eqn:gameforWGAN}
\inf_{\theta}\sup_{\substack{\rho\\
                  D_{\rho} \text{ is 1-Lipschitz}}}
                  \mathbb{E}_{x\sim P_r}[D_{\rho}(x)] - \mathbb{E}_{z\sim Q_z}[D_{\rho}(G_{\theta}(z))].
\end{equation}
The difference between weight-clipped WGANs and WGAN-GP is mainly on the technique of imposing the Lipschitz constraint for $D_{\rho}$: weight-clipped WGANs force $\rho$ to be bounded during the training, while WGAN-GP adds a gradient penalty to the loss function for the discriminator\footnote{In~\cite{arjovsky2017wasserstein} and~\cite{gulrajani2017improved}, the discriminators are names as ``critics'', but in this paper we still use the name of discriminators for consistency with other versions of GANs, including vanilla GANs.}. To be specific, in WGAN-GP the loss functions for the generator and discriminator are defined as
\begin{equation} \label{eqn:lossforWGANGP}
\begin{aligned}
	L_g &= -\mathbb{E}_{z\sim Q_z}[D_{\rho}(G_{\theta}(z))] ,\\
	L_d &= \mathbb{E}_{z\sim Q_z}[D_{\rho}(G_{\theta}(z))] - \mathbb{E}_{x\sim P_r}[D_{\rho}(x)] + \lambda \mathbb{E}_{\hat{x}\sim P_{\hat{x}} } [(\Vert\nabla_{\hat{x}} D_{\rho}(\hat{x}) \Vert_2 - 1)^2],
\end{aligned}
\end{equation}
where $P_{\hat{x}}$ is the distribution generated by uniform sampling on straight lines between pairs of points sampled from $P_r$ and $P_g$, and $\lambda$ is the gradient penalty coefficient.

Instead of the JS divergence in vanilla GANs, in WGANs the loss function for the generator corresponds to the Earth Mover or Wasserstein-1 distance ($\mathbb{W}_1$) between $P_r$ and $P_g$:
\begin{equation} \label{eqn:WD}
\mathbb{W}_1(P_r, P_g) = \inf_{\gamma \in \Gamma(P_r,P_g)} \mathbb{E}_{(x,y)\sim \gamma}[\left\lVert x-y \right\rVert],
\end{equation}
where $\Gamma(P_r,P_g)$ denotes the set of all joint distributions $\gamma(x, y)$ whose marginals are $P_r$ and $P_g$, respectively. $\mathbb{W}_1$ distance is continuous and differentiable almost everywhere with respect to the parameters in the generator under a mild constraint~\cite{arjovsky2017wasserstein}. As a result, WGANs do not suffer from the problem of mode collapse as would occur to vanilla GANs~\cite{arjovsky2017wasserstein}. Moreover, according to~\cite{gulrajani2017improved}, training WGAN-GP is more stable than weight-clipped WGANs.

\begin{figure}[H]
	\centering
	\includegraphics[width=1.0\textwidth]{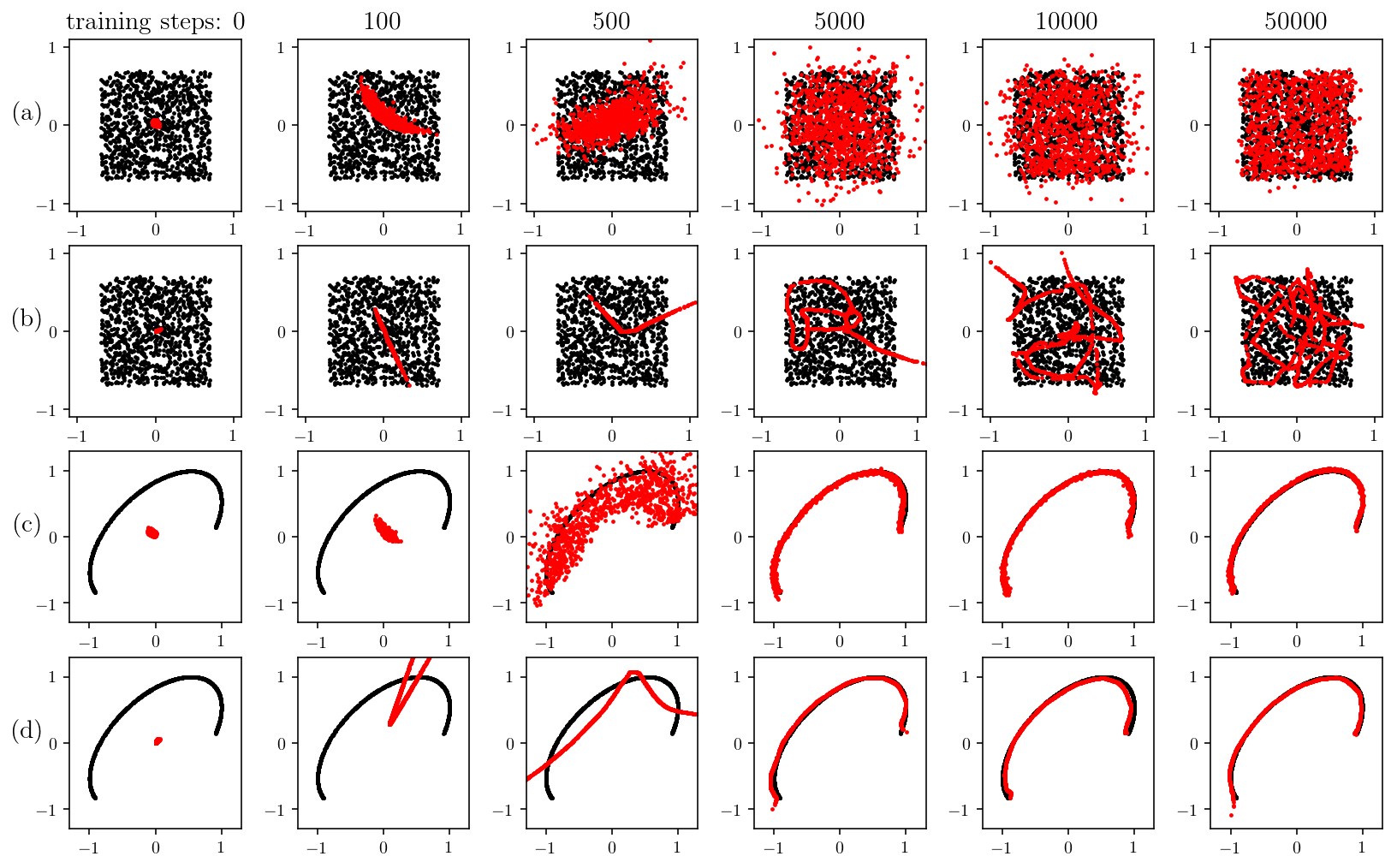}
	\caption{Evolution of generated distributions $P_g$ for two different input dimensions and two different target distributions during training. The red dots are samples from $P_g$ and the black dots are samples from $P_r$. We visualize the distributions in $\R^{10}$ by plotting sample projections in the first two dimensions. (a) The support of $P_r$ is a hypercube, $Q_z = N(0,I_{10})$. (b) The support of $P_r$ is a hypercube, $Q_z = N(0,1)$. Notice that the support of $P_g$ is always a curve, but it gets increasingly twisted in filling the hypercube and minimizing $\mathbb{W}_1(P_r, P_g)$. (c) The support of $P_r$ is a curve, $Q_z = N(0,I_{10})$. Notice that $P_g$ gets increasingly concentrated in fitting the curve and minimizing $\mathbb{W}_1(P_r, P_g)$. (d) The support of $P_r$ is a curve, $Q_z = N(0,1)$.}
	\label{fig:WGAN_Toy}
\end{figure}

To demonstrate the effectiveness of WGAN-GP, we apply it on four toy problems of approximating distributions in $\R^{10}$, as illustrated in Figure \ref{fig:WGAN_Toy}. We consider two types of $P_r$: one is a uniform distribution on a hypercube in $\R^{10}$, while the other one is a uniform distribution concentrated on a curve embedded in $\R^{10}$. For both cases of $P_r$, we test with an input of either 1-D or 10-D standard Gaussian noise, i.e., $Q_z = N(0,1)$ or $Q_z = N(0,I_{10})$. In all four cases, the generator converges and generates samples with distribution $P_g$ close to the real distribution $P_r$, even for the cases where the dimension of $Q_z$ mismatches the support of $P_r$. Due to its robustness, we use WGAN-GP as our default version of GANs in this paper.

\section{Methodology}\label{sec:Methods}

\subsection{Approximating stochastic processes with GANs} \label{subsec: stoproGANs}

As a pedagogical problem, let us first consider the problem of modeling a stochastic process $f(x;\omega)$ on the domain $D \in \R^d$ with GANs. We use the total $N$ snapshots of $f(x;\omega)$ sensor data as our training set:
\begin{equation}
    \{F(\omega^{(j)})\}_{j=1}^N = \{(f(x_i;\omega^{(j)}))_{i=1}^{n} \}_{j=1}^N,
\end{equation}
where $n$ is the number of sensors and $\{x_i\}_{i=1}^n$ are positions of the sensors. We use a feed forward DNN $\tilde{f}_{\theta}(x;\boldsymbol{\xi})$ parameterized by $\theta$ as our generator to model the stochastic process $f(x;\omega)$. The generator takes the concatenation of a noise vector $\boldsymbol{\xi} \sim Q_z$ from $\R^{m}$ and the coordinate $x$ as the input, while the output is a real number representing $\tilde{f}_{\theta}(x;\boldsymbol{\xi})$. Let
\begin{equation}
    \{\tilde{F}(\boldsymbol{\xi}^{(j)})\}_{j} = \{(\tilde{f}_{\theta}(x_i;\boldsymbol{\xi}^{(j)}))_{i=1}^{n}\}_{j}
\end{equation}
be the generated ``fake'' snapshots, where $\{\boldsymbol{\xi}^{(j)} \}_{j}$ are instances of $\boldsymbol{\xi}$ with $j$ as the index.

The discriminator $D_{\rho}(\cdot)$ is implemented as another feed forward DNN parameterized by $\rho$. The GAN strategy shall be applied by feeding both the real and generated snapshots into the discriminator, and train the generator and discriminator iteratively with the Adam optimizer~\cite{kingma2014adam} according to Equation (\ref{eqn:lossforWGANGP}). The detailed algorithm is presented in Algorithm \ref{alg:process}.

\begin{algorithm}[H]
\caption{GANs for approximating stochastic processes.}
\label{alg:process}
\begin{algorithmic}
\Require training steps $n_t$, the gradient penalty coefficient $\lambda$, the number of discriminator iterations per generator iteration $n_d$, the batch size $n$, Adam hyperparameters $\alpha, \beta_1, \beta_2$, initial values $\theta_0$ and $\rho_0$ for the parameters generator $\theta$ and $\rho$.
\For{ $s_t$ = 1,2,...,$n_t$}
	\For{ $s_d$ = 1,2,...,$n_d$}
	\State Sample $n$ snapshots $\{F^{(j)}\}_{j=1}^n$ from the training data set.
	\State Sample $n$ random vectors $\{\boldsymbol{\xi}^{(j)}\}_{j=1}^n\sim Q_z$.
	\State Sample $n$ uniform random numbers $\{\epsilon^{(j)}\}_{j=1}^n \sim U[0, 1]$.
	    \For{$j$ = 1,2,...,$n$}
	        \State $\hat{F}^{(j)} \leftarrow \epsilon^{(j)} F^{(j)} + (1-\epsilon^{(j)})\tilde{F}(\boldsymbol{\xi}^{(j)})$
	        \State $L^{(j)} \leftarrow D_\rho(\tilde{F}(\boldsymbol{\xi}^{(j)})) -  D_\rho(F^{(j)}) + \lambda (\Vert\nabla_{\hat{F}^{(j)}} D_{\rho}(\hat{F}^{(j)}) \Vert_2 - 1)^2$
        \EndFor
    \State $\rho \leftarrow \text{Adam} (\nabla_{\rho} \frac{1}{n} \sum_{j=1}^n L^{(j)}, \rho, \alpha, \beta_1,\beta_2)$
    \EndFor
    \State Sample $n$ random vectors $\{\boldsymbol{\xi}^{(j)}\}_{j=1}^{n} \sim Q_z$.
    \State $\theta \leftarrow \text{Adam}(\nabla_{\theta}\frac{1}{n}\sum_{j=1}^{n} -D_{\rho}(\tilde{F}(\boldsymbol{\xi}^{(j)})), \theta, \alpha, \beta_1, \beta_2 )$
\EndFor
\end{algorithmic}
\end{algorithm}

\subsection{Solving stochastic differential equations with PI-GANs}
\begin{figure}[H]
	\centering
	\includegraphics[width=0.7\textwidth]{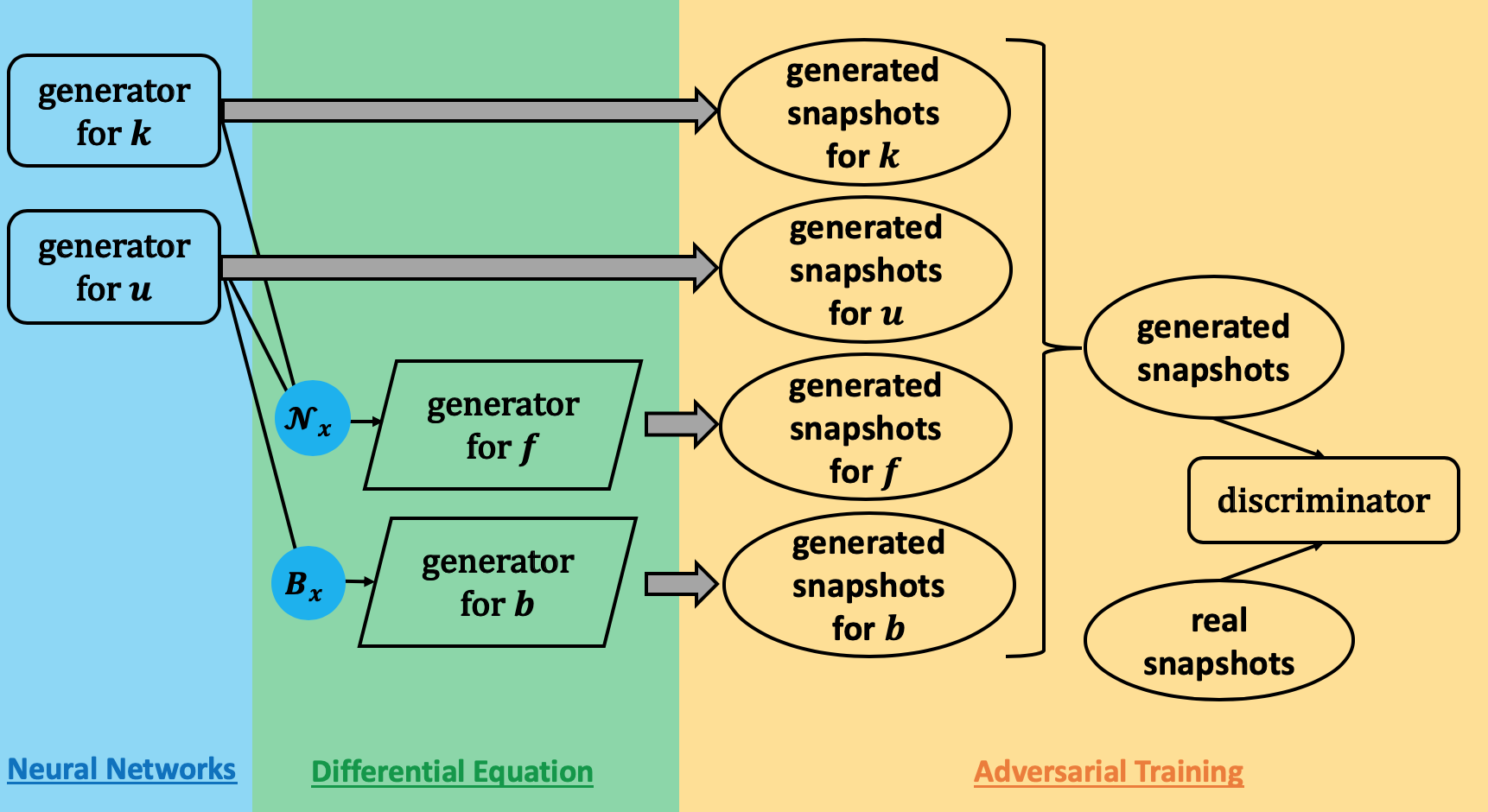}
	\caption{Schematic of solving SDEs with PI-GANs. The round corner rectangles in the blue and yellow boxes represent feed forward neural networks. The parallelograms in the green box represent the neural networks induced by operators ($\mathcal{N}_x$ and $B_x$). The ellipses represent snapshots from sensors, and the gray arrows represent sampling procedures. The bracket in the yellow box represents concatenation.}
	\label{fig:SDE2NN}
\end{figure}

Consider Equation (\ref{eqn:General}), as illustrated in Figure \ref{fig:SDE2NN}, solving SDEs with PI-GANs consists of the following three steps.

First, we use two independent feed forward DNNs, namely $\tilde{k}_{\theta_k}(x;\boldsymbol{\xi} )$ and $\tilde{u}_{\theta_u}(x;\boldsymbol{\xi} )$ parameterized by $\theta_k$ and $\theta_u$, to represent the stochastic processes $k(x;\omega)$ and $u(x;\omega)$ in the aforementioned way.

Second, inspired by the physics-informed neural networks for deterministic PDEs~\cite{MaziarParisGK17_1}, we encode the equation into the neural networks system by applying the operator $\mathcal{N}_x$ and $B_x$ on the feed forward DNNs $\tilde{k}_{\theta_k}(x;\boldsymbol{\xi} )$ and $\tilde{u}_{\theta_u}(x;\boldsymbol{\xi} )$ to generate ``induced'' neural networks, which are formulated as
\begin{equation} 
\tilde{f}_{\theta_u, \theta_k}(x;\boldsymbol{\xi}) = \mathcal{N}_x [\tilde{u}_{\theta_u}(x;\boldsymbol{\xi}); \tilde{k}_{\theta_k}(x;\boldsymbol{\xi})]
\end{equation}
and 
\begin{equation} 
\tilde{b}_{\theta_u}(x;\boldsymbol{\xi}) =B_x[\tilde{u}_{\theta_u}(x;\boldsymbol{\xi})].
\end{equation}
Differentiation in $\mathcal{N}_x$ and $B_x$ are performed by automatic differentiation~\cite{baydin2017automatic}. We then use $\tilde{f}_{\theta_u, \theta_k}(x;\boldsymbol{\xi})$ and $\tilde{b}_{\theta_u}(x;\boldsymbol{\xi})$ as the generators of $f(x;\omega)$ and $b(x;\omega)$, respectively. Note that both $\tilde{f}_{\theta_u, \theta_k}(x;\boldsymbol{\xi})$ and $\tilde{b}_{\theta_u}(x;\boldsymbol{\xi})$ are induced from $\tilde{u}_{\theta_u}(x;\boldsymbol{\xi})$ and $\tilde{k}_{\theta_k}(x;\boldsymbol{\xi})$, thus the parameters $\theta_k$ and $\theta_u$ are directly inherited from $\tilde{u}$ and $\tilde{k}$.

In the third step, we incorporate our training data to conduct adversarial training. The training data are collected as a group of snapshots in Equation (\ref{eqn:Data}). The corresponding generated ``fake'' snapshots are
\begin{equation} \label{eqn:G}
\begin{aligned}
\{G(\boldsymbol{\xi}^{(j)})\}_j & = \{(\tilde{K}(\boldsymbol{\xi}^{(j)}), \tilde{U}(\boldsymbol{\xi}^{(j)}),\tilde{F}(\boldsymbol{\xi}^{(j)}),\tilde{B}(\boldsymbol{\xi}^{(j)}))\}_j,\\
\tilde{K}(\boldsymbol{\xi}^{(j)}) & = (\tilde{k}_{\theta_k}(x_i^k;\boldsymbol{\xi}^{(j)}))_{i=1}^{n_k},\\
\tilde{U}(\boldsymbol{\xi}^{(j)}) & = (\tilde{u}_{\theta_u}(x_i^u;\boldsymbol{\xi}^{(j)}))_{i=1}^{n_u},\\
\tilde{F}(\boldsymbol{\xi}^{(j)}) & = (\tilde{f}_{\theta_k,\theta_u}(x_i^f;\boldsymbol{\xi}^{(j)}))_{i=1}^{n_f},\\
\tilde{B}(\boldsymbol{\xi}^{(j)}) & = (\tilde{b}_{\theta_u}(x_i^b;\boldsymbol{\xi}^{(j)}))_{i=1}^{n_b},
\end{aligned}
\end{equation}
where $\{\boldsymbol{\xi}^{(j)} \}_{j}$ are instances of $\boldsymbol{\xi}$ with $j$ as the index. We could then feed the generated snapshots and real snapshots into discriminator, and train the generators and discriminators iteratively. With well trained generators, we can then calculate all the statistics with sample paths created by the generators.

If the snapshots are collected in $M$ groups ($M>1$), we will also need to generate $M$ groups of ``fake'' snapshots:
\begin{equation} \label{eqn:G_multi}
\begin{aligned}
\{\{G_t(\boldsymbol{\xi}^{(t,j)}\}_j\}_{t=1}^M & = \{\{(\tilde{K_t}(\boldsymbol{\xi}^{(t,j)}), \tilde{U_t}(\boldsymbol{\xi}^{(t,j)}),\tilde{F_t}(\boldsymbol{\xi}^{(t,j)}),\tilde{B_t}(\boldsymbol{\xi}^{(t,j)}))\}_j\}_{t=1}^{M},\\
\tilde{K_t}(\boldsymbol{\xi}^{(t,j)}) & = (\tilde{k}_{\theta_k}(x_i^{k,t};\boldsymbol{\xi}^{(t,j)}))_{i=1}^{n_{k,t}},\\
\tilde{U_t}(\boldsymbol{\xi}^{(t,j)}) & = (\tilde{u}_{\theta_u}(x_i^{u,t};\boldsymbol{\xi}^{(t,j)}))_{i=1}^{n_{u,t}},\\
\tilde{F_t}(\boldsymbol{\xi}^{(t,j)}) & = (\tilde{f}_{\theta_k,\theta_u}(x_i^{f,t};\boldsymbol{\xi}^{(t,j)}))_{i=1}^{n_{f,t}},\\
\tilde{B_t}(\boldsymbol{\xi}^{(t,j)}) & = (\tilde{b}_{\theta_u}(x_i^{b,t};\boldsymbol{\xi}^{(t,j)}))_{i=1}^{n_{b,t}},\\
\end{aligned}
\end{equation}
where $t$s are the indices for the groups, $\boldsymbol{\xi}^{(t,j)}$ is an instance of $\boldsymbol{\xi}$ for each $(t,j)$, $\{x^{k,t}_i\}_{i=1}^{n_{k,t}}$ is the position setup of $n_{k,t}$ sensors for $k$ in group $t$ (similarly for other terms). 
We will also use multiple discriminators $\{D_{\rho_t}(\cdot)\}_t^M$, with each discriminator focusing on one group of snapshots, while the generators need to ``deceive'' all the discriminators simultaneously. 

We give a formal and detailed description of our method in Algorithm \ref{alg:SDE_multi}. For the case where we only have one group of data, we set $M = 1$. For simplicity, here we set the weight in the generator loss function $a_t = 1$ for each $t$, which works well, but the method of setting $\{a_t\}_{t=1}^M$ requires further study.

\begin{algorithm}[H]
\caption{PI-GANs for solving stochastic differential equations.}
\label{alg:SDE_multi}
\begin{algorithmic}
\Require training steps $n_t$, the gradient penalty coefficient $\lambda$, the number of discriminator iterations per generator iteration $n_d$, the batch size $n$, Adam hyper-parameters $\alpha, \beta_1, \beta_2$, initial values ${\theta_k}_0$, ${\theta_u}_0$ and $\{{\rho_t}_0\}_{t=1}^M$ for the parameters ${\theta_k}$, ${\theta_u}$ and $\{\rho_t\}_{t=1}^M$, the weights in the loss function for the generators $\{a_t\}_{t=1}^M$
\For{ $s_t$ = 1,2,...,$n_t$}
	\For{ $s_d$ = 1,2,...,$n_d$}
	    \For{$t$= 1,2,...,$M$}
	\State Sample $n$ snapshots $\{T_t^{(j)}\}_{j=1}^n$ from training data group $t$.
	\State Sample $n$ random vectors $\{\boldsymbol{\xi}^{(t,j)}\}_{j=1}^n\sim Q_z$.
	\State Sample $n$ uniform random numbers $\{\epsilon^{(t,j)}\}_{j=1}^n \sim U[0, 1]$.
	    \For{$j$ = 1,2,...,$n$}
	        \State $\hat{G_t}^{(j)} \leftarrow \epsilon^{(j)} T_t^{(j)} + (1-\epsilon^{(j)})G_t(\boldsymbol{\xi}^{(t,j)})$
	        \State $L_t^{(j)} \leftarrow D_{\rho_t}(G_t(\boldsymbol{\xi}^{(t,j)})) -  D_{\rho_t}(T^{(t,j)})\newline  
	        \hspace*{9em}+ \lambda (\Vert\nabla_{\hat{G_t}^{(j)}} D_{\rho_t}(\hat{G_t}^{(j)}) \Vert_2 - 1)^2$
        \EndFor
    \State $\rho_t \leftarrow \text{Adam} (\nabla_{\rho_t} \frac{1}{n} \sum_{j=1}^n L_t^{(j)}, \rho_t, \alpha, \beta_1,\beta_2)$
    \EndFor
    \EndFor
    \State Sample $n$ random vectors $\{\boldsymbol{\xi}^{(j)}\}_{j=1}^{n} \sim Q_z$.
    \State $\theta \leftarrow \text{Adam}(\nabla_{\theta}\sum_{t=1}^M {a_t}(\frac{1}{n}\sum_{j=1}^{n} -D_{\rho_t}(G_t(\boldsymbol{\xi}^{(j)}))), \theta, \alpha, \beta_1, \beta_2 )$, \newline 
    \hspace*{3em} where $\theta = (\theta_k, \theta_u)$
\EndFor
\end{algorithmic}
\end{algorithm}

Note that our method does not explicitly distinguish the three types of problems described Section \ref{S:1.5}. Solving forward problems, inverse problems or mixed problems actually uses the same framework.
 
\section{Numerical Results}\label{sec:results}
The following settings are commonly shared by all the test cases. We use \textit{tanh} as the activation function instead of the commonly used ReLU activation function, because piece-wise linear functions are not suitable for solving SDEs, where we may need to take high order derivatives. All the feed forward DNNs for generators in the following numerical experiments have 4 hidden layers of width 128. The sizes of the input layer into the generators vary and will be specified case by case. Most of the discriminators also have hidden layers of the same size as the generators, except the ones in Section \ref{sec:Toy} that have 4 hidden layers of width 64, and the one for the additional group of snapshots on $u(x;\omega)$ in Section \ref{sec:Inv} that has 4 hidden layers of width 16. To initialize the DNNs, we use the uniform Xavier initializer for weights and zero initializer for biases. The distribution $Q_z$ of the noise input into the generators is an independent multi-variate standard Gaussian distribution. For the hyper-parameters in loss functions and optimizers, we use the default values of $\lambda = 0.1, n_d = 5, \alpha = 0.0001, \beta_1 = 0, \beta_2 = 0.9$, as in the toy problem in~\cite{gulrajani2017improved}. The sensors are placed equidistantly in the domain. Our algorithms are implemented with Tensorflow.

\subsection{A pedagogical problem: approximating stochastic processes} \label{sec:Toy}
In this section, we test the problem of approximating stochastic processes. Consider the following Gaussian processes with zero mean and squared exponential kernel:
\begin{equation} \label{eqn:GP}
\begin{aligned}
    f(x) & \sim \mathcal{GP}\left(0, \exp(\frac{-(x-x')^2}{2l^2})\right),\\
    x &\in D =  [-1,1],
\end{aligned}
\end{equation}
where $l$ is the correlation length.

\subsubsection{Effect of the number of sensors and snapshots}
\begin{figure}[H]
    \centering
	\includegraphics[width=1.0\textwidth]{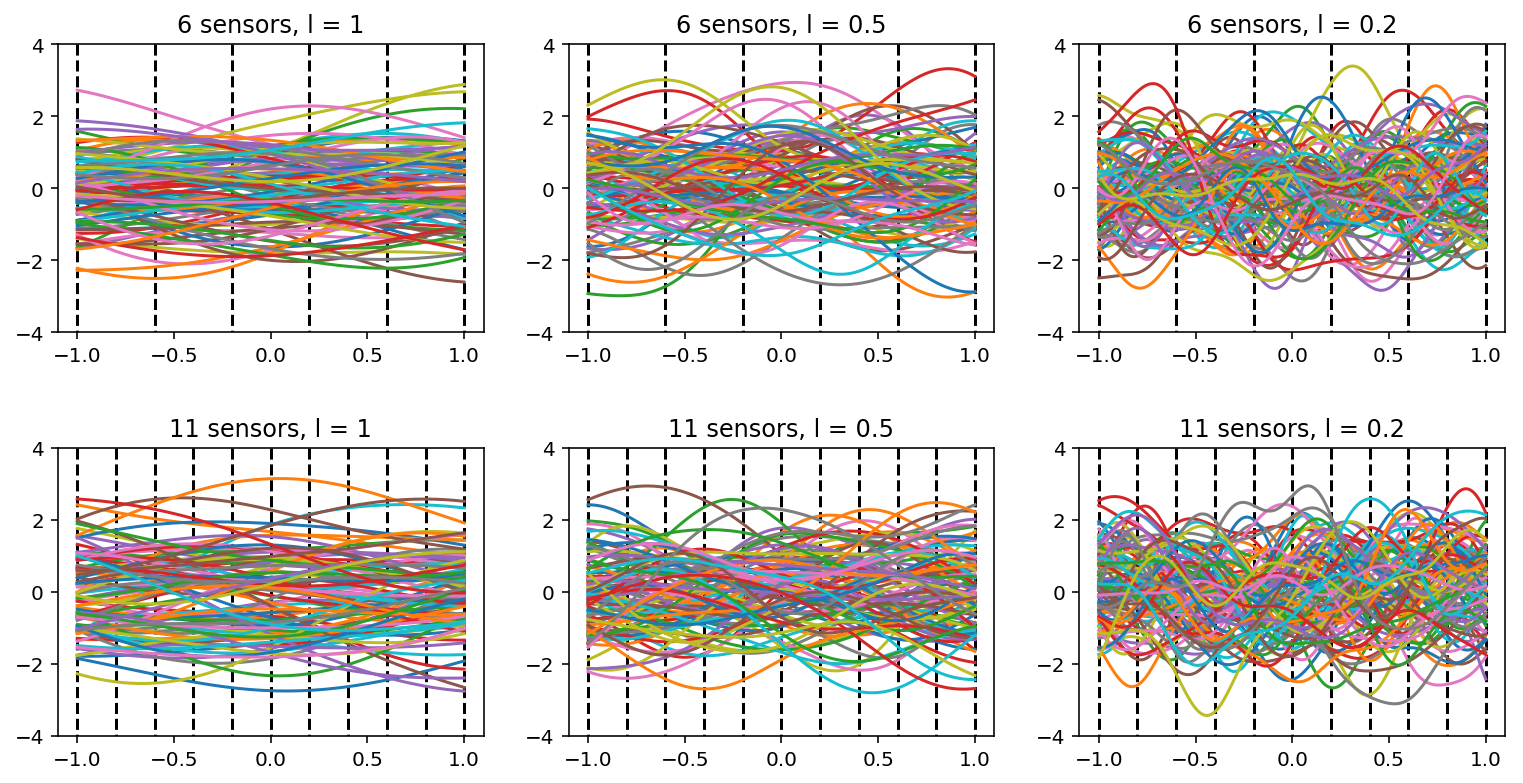}
	\caption{Sample paths of Gaussian processes with the sensor locations denoted by the vertical dashed lines. We use the correlation length $l = $ 1 (left), 0.5 (middle), 0.2 (right), and 6 sensors (top) and 11 sensors (bottom).}
	\label{fig:0_samples}
\end{figure}

We consider three different choices of correlation length $l$: $l = 1, 0.5, 0.2$, and two sensor numbers for $f(x;\omega)$: 11 and 6, and fix the number of snapshots to be 1000. For $l = 0.2$ we also consider a supplementary case, where we have $1\times10^4$ snapshots. The training sample paths and positions of sensors are illustrated in Figure \ref{fig:0_samples}. For each case, we run the code three times with different random seeds. The batch size in all the cases is 1000. The input layer of the generators has width of 5, i.e., the input noise is a four-dimensional random vector.

\begin{figure}[H]
    \centering
	\includegraphics[width=0.6\textwidth]{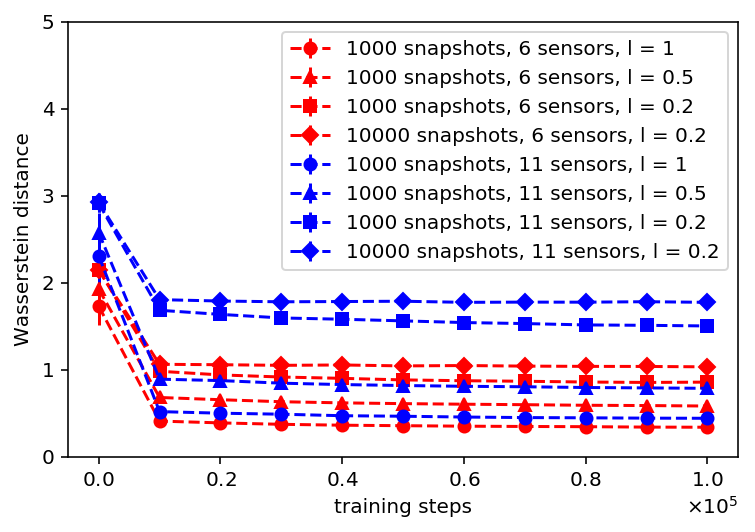}
	\caption{$\mathbb{W}_1$ distances $\mathbb{W}_1(\hat{P^{n}_g},\hat{P^{n}_t})$ versus training steps, for different correlation lengths and number of sensors. The decay of $\mathbb{W}_1$ distances becomes slow after about $2\times10^4$ steps and hard to see after $8\times10^4$ steps. We set the number of sample snapshots for empirical distribution to be $n = 1000$. The means and two standard deviations are from 3 independent runs and 10 batches of $n$ generated snapshots and training snapshots in each run.
	}
	\label{fig:0_Wass}
\end{figure}

To decide when to stop the training process, we calculate the $\mathbb{W}_1$ distances between the empirical distributions of generated snapshots and training snapshots $\mathbb{W}_1(\hat{P^n_g},\hat{P^n_t})$, where $\hat{P^n_g}$ is the empirical distribution of $n$ generated snapshots from Monte Carlo sampling and $\hat{P^n_t}$ is the empirical distribution of $n$ snapshots from the training set, using the python POT package \cite{flamary2017pot}. In our tests, we set $n=1000$. We plot the $\mathbb{W}_1$ distances in Figure \ref{fig:0_Wass}. Note that the $\mathbb{W}_1$ distances decay and converge during the training, indicating that the generated distributions approach the real distributions. We stop the training after $1\times 10^5$ steps since $\mathbb{W}_1(\hat{P^n_g},\hat{P^n_t})$ is stable. In each run, we select the 11 generators at training step in the last 10001 steps with a stride of $1\times10^3$; all together, we have 33 generators for each case. From each selected generator, we generate $1\times 10^4$ sample paths based on the Halton quasi-Monte Carlo method, and calculate its spectra, i.e. the eigenvalues of the covariance matrices from the principal component analysis.

\begin{figure}[H]
	\centering
	\includegraphics[width=1.0\textwidth]{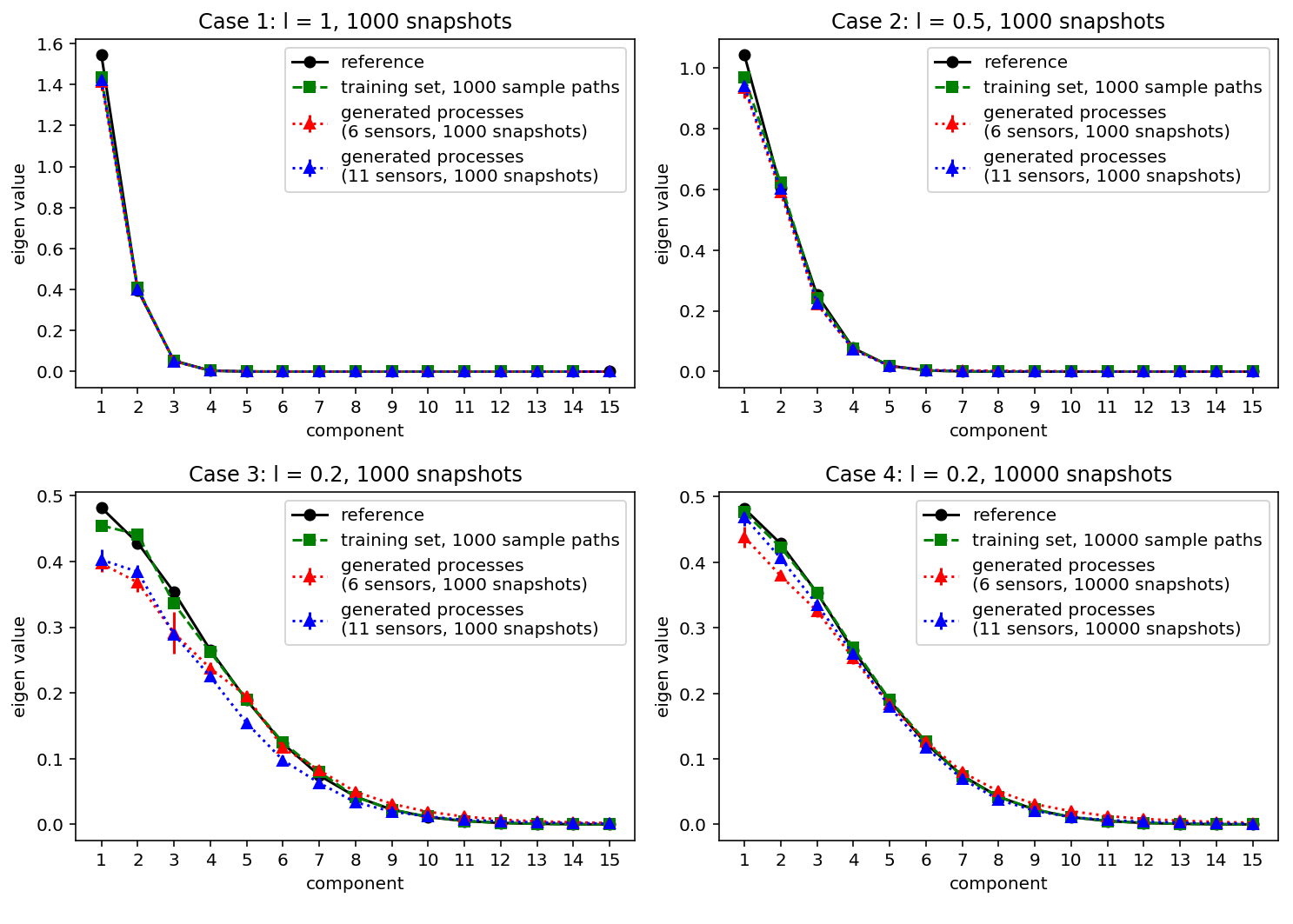}
	\caption{Spectra of the correlation structure for the generated processes for different correlation lengths. The training sets consist of the processes where the training snapshots are collected from, and the reference is calculated from another $1\times 10^5$ independent sample paths. The means and two standard deviations are calculated from the selected 33 generators for each case; the standard deviations are very small.}
	\label{fig:0_PCA}
\end{figure}


The results are illustrated in Figure \ref{fig:0_PCA}, from which we conclude that: 
\begin{enumerate}
    \item When we fix the number of snapshots to be 1000, as the correlation length decreases, the gap between our generated processes and the reference solutions becomes wider. This is because smaller $l$ results in higher effective dimension, and more subtle local behavior of the stochastic processes; however, this gap could be narrowed if we increase the number of snapshots.
    \item The approximations in the cases with 11 sensors are better than the approximations in the cases with 6 sensors if we have sufficient training data. This is reasonable since we need more sensors to describe the stochastic processes with small correlation length. 
    \item Despite the fact that the input noise into the generator is a four-dimensional vector, the spectra of the generated processes approximate the spectra of the target processes with much higher effective dimensionality. This makes sense since the low dimensional manifold could fold and twist itself to fill the high dimensional region, as discussed earlier.
\end{enumerate}

\subsubsection{WGAN-GP versus vanilla GANs}
We compare the performance of WGAN-GP and the vanilla GANs in approximating stochastic processes for the following two cases: 
\begin{enumerate}
    \item A Gaussian process in Equation (\ref{eqn:GP}) with $l=0.2$.
    \item A stochastic process with fixed 0 boundary condition. Specifically, we consider
    \begin{equation} \label{eqn:bound_GP}
    \begin{aligned}
    f(x) & = (x^2 - 1) g(x), \\
    g(x) & \sim \mathcal{GP}(0, \exp(\frac{-(x-x')^2}{2 \times 0.2^2})), \\
    x &\in D =  [-1,1]. \\
    \end{aligned}
    \end{equation}
\end{enumerate}

\begin{figure}[H]
	\centering
	\includegraphics[width=1.0\textwidth]{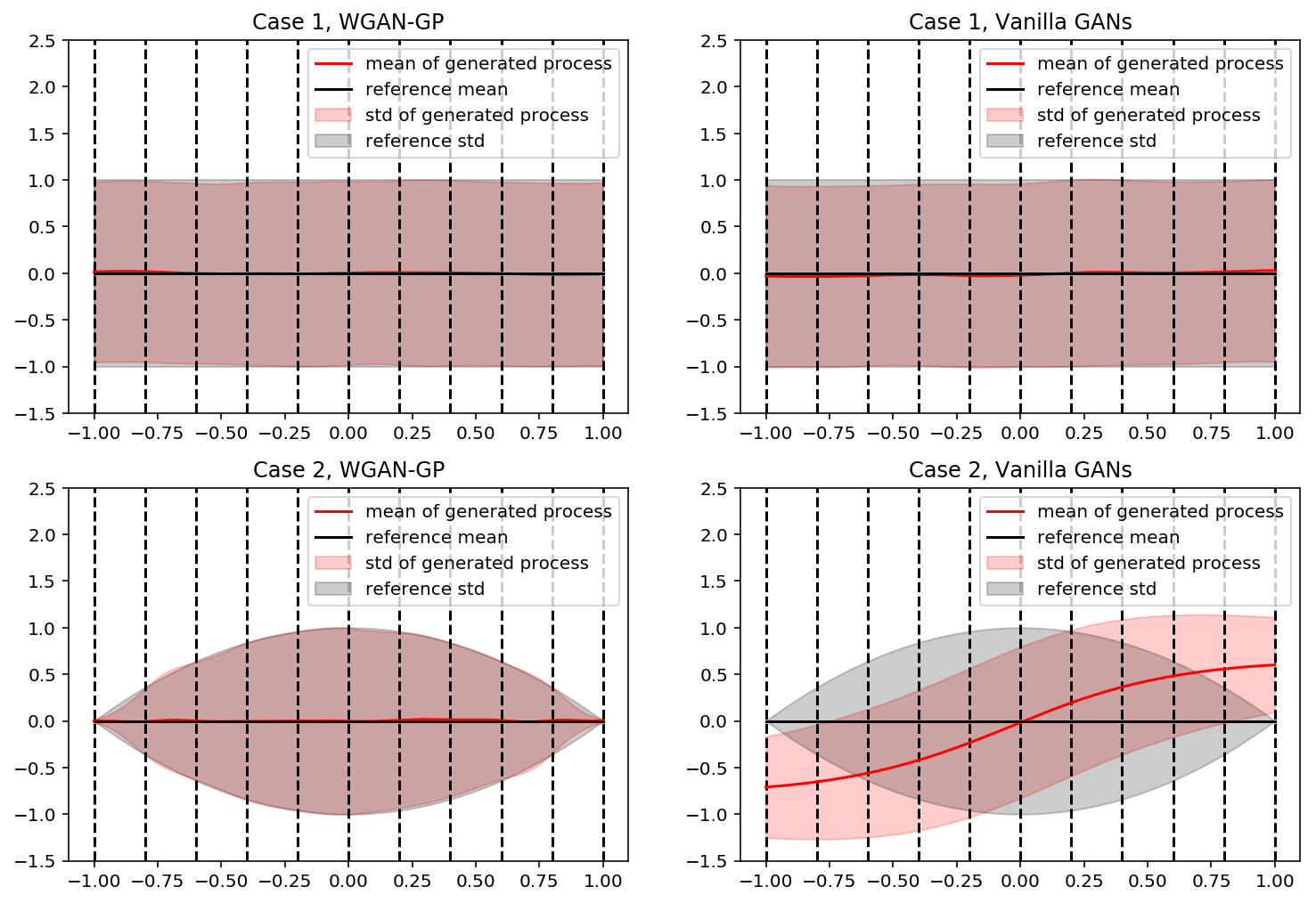}
	\caption{Mean and standard deviation of random processes generated by WGAN-GP and vanilla GANs: For case 1, both versions of GANs perform well, but the vanilla GANs failed in case 2, while the WGAN-GP can still provide accurate results. The vertical dashed lines represent the positions of sensors.}
	\label{fig:0_compare}
\end{figure}

In both cases, we use $1\times10^4$ snapshots collected from 11 sensors as training data. In Figure \ref{fig:0_compare} we show the means and standard deviations of generated processes trained by WGAN-GP and the vanilla GANs. We can see that both versions of GANs produce good approximations for case 1. However, the vanilla GANs fail in case 2, while WGAN-GP still generates a good approximation. This agrees with the theory in~\cite{gulrajani2017improved} that vanilla GANs are not suitable for approximating distributions concentrated on low dimensional manifolds (the two fixed boundaries in this case), since the JS divergence cannot provide a usable gradient for the generators. 

\subsubsection{Overfitting issues}
\begin{figure}[H]
	\centering
	\begin{subfigure}[b]{0.5\textwidth}
	\centering
	\includegraphics[width=1.0\textwidth]{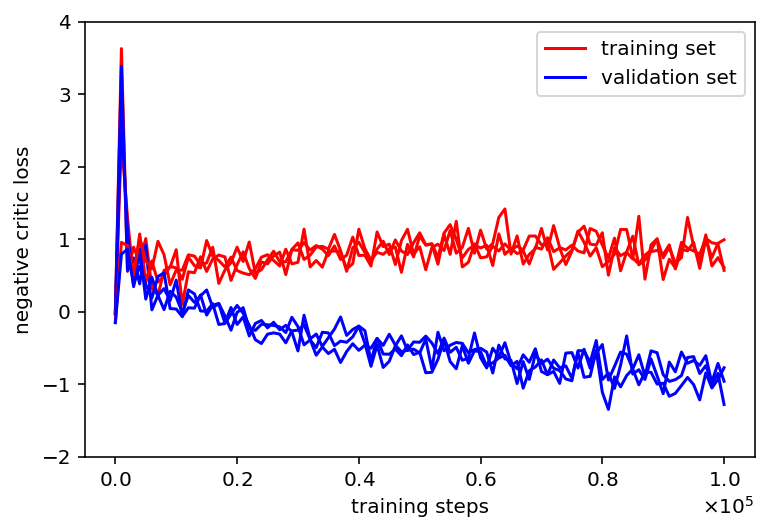}
	\caption{}
	\label{fig:4_loss}
	\end{subfigure}%
	\begin{subfigure}[b]{0.5\textwidth}
	\centering
	\includegraphics[width=1.0\textwidth]{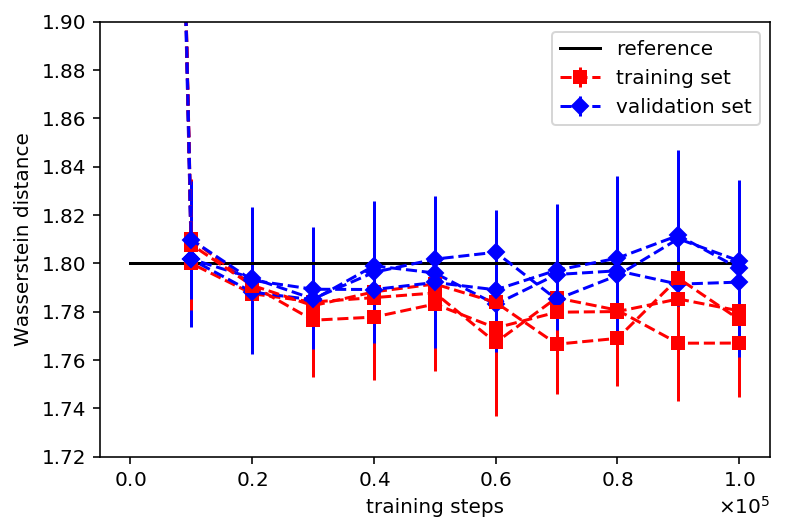}
	\caption{}
	\label{fig:4_OT}
	\end{subfigure}
	\caption{Overfitting of the discriminator {\em and} generator. (a) Negative discriminator loss on the training set and validation set versus the training steps in three independent runs. (b) $\mathbb{W}_1$ distances $\mathbb{W}_1(\hat{P^{n}_g},\hat{P^{n}_t})$ and $\mathbb{W}_1(\hat{P^{n}_g},\hat{P^{n}_v})$ versus the training steps in three independent runs. The means (markers) and two standard deviations (vertical lines on the marker) come from ten groups of $(\hat{P^{n}_g}, \hat{P^{n}_t}, \hat{P^{n}_v})$. The black horizontal line is the empirical expectation of $\mathbb{W}_1(\hat{P^{n}_r}_1,\hat{P^{n}_r}_2)$ from 50 groups of $(\hat{P^{n}_r}_1,\hat{P^{n}_r}_2)$.}
\end{figure}

It was pointed out in \cite{gulrajani2017improved} that the discriminator can overfit the training data given sufficient capacity but too little training data. Here we report the same issue in our method. Take the case of approximating the stochastic process where the correlation length is $l=0.2$ with 11 sensors and 10000 training snapshots as an example. We plot the negative discriminator loss $-L_d$ for the training set and the validation set in Figure \ref{fig:4_loss}. As training goes on, the negative discriminator loss gradually increases for the training set while still decreases for the validation set. The gap between them implies that the discriminator overfits the training data, and gives a biased estimation of the $\mathbb{W}_1$ distance between the real distribution and generated distribution, just as is reported in~\cite{gulrajani2017improved}.

How about the overfitting of generators? In our problem, the overfitting of generators comes in two types:
\begin{enumerate}
    \item[Type-1]: Overfitting in the random space: The distribution of generated snapshots is biased towards the empirical distribution of the training snapshots. In the worst case, the generated snapshots are concentrated on or near the support of training data.
    \item[Type-2]: Overfitting in the physical space: The generated stochastic processes become worse after extensive training, and tend to match the real processes only at the sensor locations and display large variations where there is no sensor. 
\end{enumerate}

We first report that we did not detect type-2 overfitting in our experiments. Actually, this type of overfitting would be reflected on the mean and standard deviation of generated processes, which fit the reference values pretty well in our experiments. We attribute this to the property of GANs that the target for the generator is to approximate a distribution rather than a single point on the sensors. As a result, the generator does not need to overfit a specific value on the sensors in order to decrease the loss. 

As for the type-1 overfitting, we could detect it in our experiments. As depicted in Figure \ref{fig:4_OT}, we can see this from the $\mathbb{W}_1$ distances between empirical distributions of the generated snapshots and the training snapshots or the validation snapshots, i.e., $\mathbb{W}_1(\hat{P^{n}_g},\hat{P^{n}_t})$ or $\mathbb{W}_1(\hat{P^{n}_g},\hat{P^{n}_v})$, where $\hat{P^{n}_g}$, $\hat{P^{n}_t}$ and $\hat{P^{n}_v}$ are empirical distributions of generated snapshots, training snapshots and validation snapshots, $n$ is the number of snapshots. Here, we set $n =1000$. We can see that as the training goes on, $\mathbb{W}_1(\hat{P^{n}_g},\hat{P^{n}_v})$ converges around the expectation of $\mathbb{W}_1(\hat{P^{n}_r}_1,\hat{P^{n}_r}_2)$, where $\hat{P^{n}_r}_1$ and $\hat{P^{n}_r}_2$ are independent empirical distributions of $n$ snapshots from real distribution $P_r$. However, $\mathbb{W}_1(\hat{P^{n}_g},\hat{P^{n}_t})$ goes down below the reference line, indicating that the generated snapshots are biased towards the training snapshots, in other words, type-1 overfitting actually happened.


Finally, we point out that type-1 overfitting is less harmful than underfitting: in our problems, even in the worst case where the generated distribution is concentrated on or near the support of training data, we can still recover the sample paths whose snapshots are concentrated on or near the training snapshots, and based on these sample paths we can still obtain decent estimations. 

\subsection{Forward problem} \label{sec:Forward}
\subsubsection{Case 1: Effects of input noise dimension and number of training snapshots}
\begin{figure}[H]
	\centering
	\includegraphics[width=1.0\textwidth]{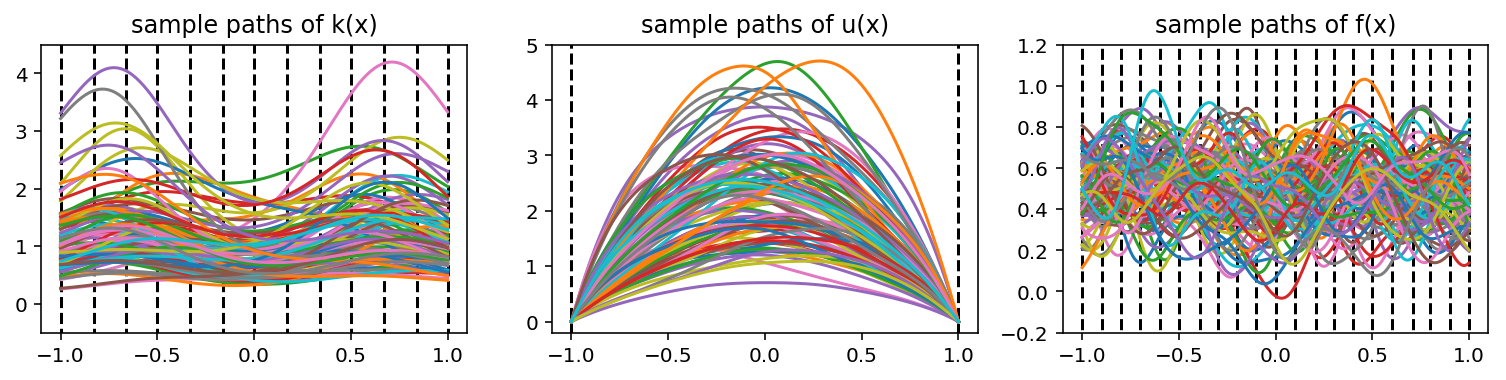}
	\caption{Forward problem: the sample paths and sensor locations of $k$, $u$, and $f$. The vertical dashed lines denote the locations of sensors.
}
	\label{fig:1_samplepaths}
\end{figure}

We consider the stochastic elliptic equation (Equation (\ref{eqn:kufequation})), and $k(x;\omega)$ and $f(x;\omega)$ as the following independent stochastic processes:
\begin{equation} \label{eqn:samples}
\begin{aligned}
	 k(x) & = \exp[\frac{1}{5}\sin(\frac{3\pi}{2}(x +1)) + \hat{k}(x)] \\
	\hat{k}(x) & \sim \mathcal{GP}\left(0, \frac{4}{25}\exp(-(x-x')^2)\right) \\
	f(x) & \sim \mathcal{GP}\left(\frac{1}{2}, \frac{9}{400}\exp(-25(x-x')^2)\right) \\
\end{aligned}
\end{equation}
In this case, we need 13 dimensions to retain $99\%$ energy of $f(x;\omega)$. We put 13 $k$-sensors , 21 $f$-sensors and 2 $u$-sensors on the boundary of physical domain $\mathcal{D}$. The positions of the sensors and some sample paths of $k(x;\omega)$, $u(x;\omega)$ and $f(x;\omega)$ are illustrated in Figure \ref{fig:1_samplepaths}.

To study the influence of input noise dimension, we fix the number of training snapshots to be 1000, and vary the input noise dimension to be 2, 4, 20 and 50. Subsequently, we fix the input noise dimension as 20 and vary the number of training snapshots to be 300, 1000 and 3000 to study the influence of the number of training snapshots. During the training process, we keep the batch size to be the total number of training snapshots. One notable condition here is the independence of $k(x;\omega)$ and $f(x;\omega)$. To reflect this, we shuffle the alignment of snapshots from $k(x;\omega)$ and $f(x;\omega)$ in each training step. For each case, we run the code three times with different random seeds. We stop the training after $1\times 10^5$ steps, and then select 33 generators and generate $1\times 10^4$ sample paths from each generator in the same way as in Section \ref{sec:Toy} to calculate the following statistics.

\begin{figure}[H]
	\centering
	\includegraphics[width=1.0\textwidth]{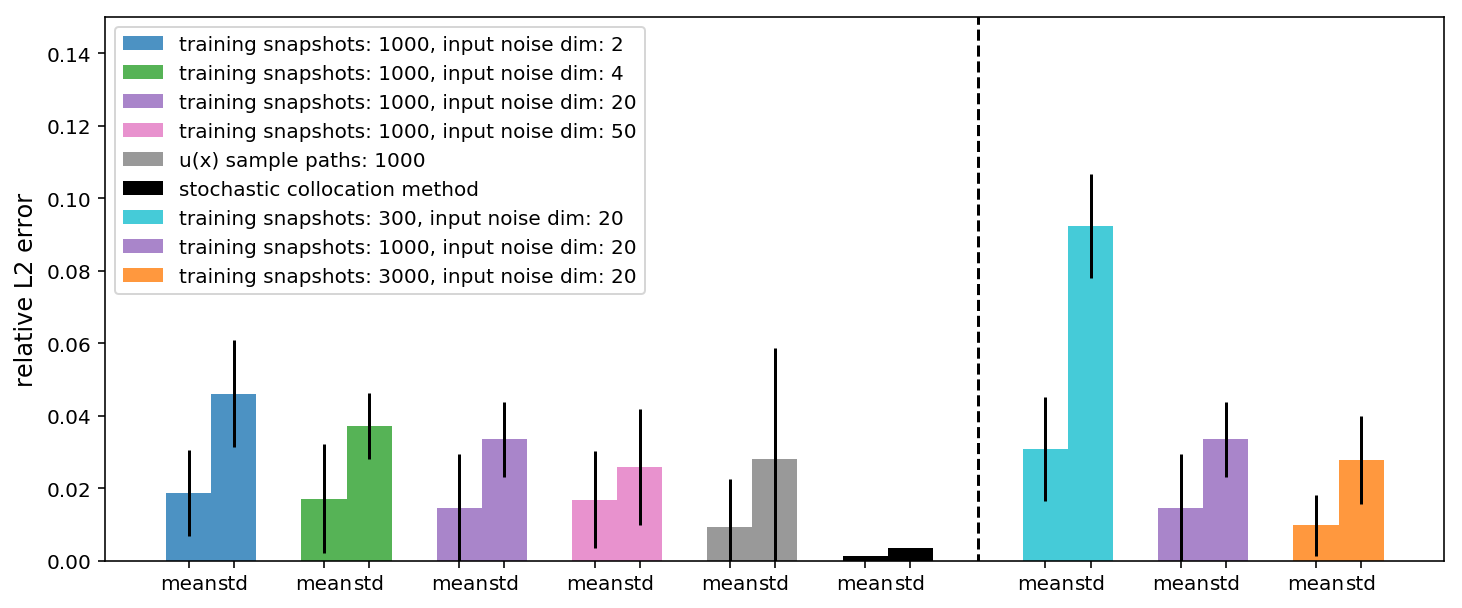}
	\caption{Forward problem for different input noise dimensionality (left part of the panel) and the number of training snapshots (right part of the panel): relative errors of inferred mean and standard deviation of the stochastic solution $u(x;\omega)$. The colored bars and the corresponding black lines represent the mean and two standard deviations of the relative errors calculated from the selected 33 generators for each case. The grey bars and the black lines represent the expectation and two standard deviations of the relative error if we calculate the mean and standard deviation of $u(x;\omega)$ from a random draw of 1000 $u(x;\omega)$ sample paths. The black bar (before the vertical dashed line) is the relative error from stochastic collocation method with 630 collocation points. The reference solutions are calculated from $1 \times 10^5$ $u(x;\omega)$ sample paths.}
	\label{fig:1_MeanError_all}
\end{figure}

Our main quantity of interest in this problem is the mean and standard deviation of $u(x;\omega)$. In Figure \ref{fig:1_MeanError_all} we show the relative error of our inferences, and compare it with the relative error of the stochastic collocation method and the Monte Carlo full trajectory sample paths of $u(x;\omega)$. We can see that:
\begin{enumerate}
    \item Our errors are comparable with the errors calculated from 1000 full trajectory sample paths of $u(x;\omega)$, showing the effectiveness of our method considering that we only have 1000 snapshots on sparsely placed sensors for $k(x;\omega)$, $f(x;\omega)$ and boundary of $u(x;\omega)$.
    \item The stochastic collocation method gives a better solution, but it requires a full knowledge of $k(x;\omega)$ and $f(x;\omega)$, including the covariance kernel function, which is far beyond our accessible data. 
    \item When we increase the dimension of input noise, we can see that the error of mean does not change too much, but the error of the standard deviation decreases. We attribute this to the fact that although a low dimensional manifolds could twist itself to fill in the high dimensional region, higher dimensional manifolds produce better approximations by filling in the high dimensional region more efficiently.
    \item With fixed input noise dimension, the error decreases as the number of training snapshots increases.
\end{enumerate}

\begin{figure}[H]
	\centering
	\begin{subfigure}[b]{0.5\textwidth}
	    \centering
	    \includegraphics[width=1.0\textwidth]{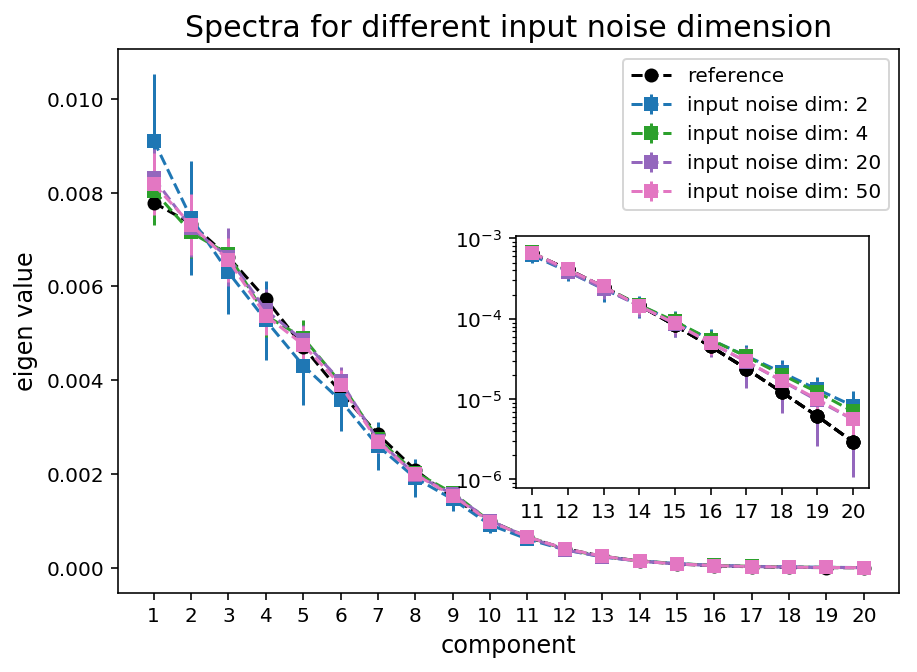}	
	    \caption{}
	    \label{fig:1_PCA_1}
	\end{subfigure}%
	\begin{subfigure}[b]{0.5\textwidth}
	    \centering
	    \includegraphics[width=1.0\textwidth]{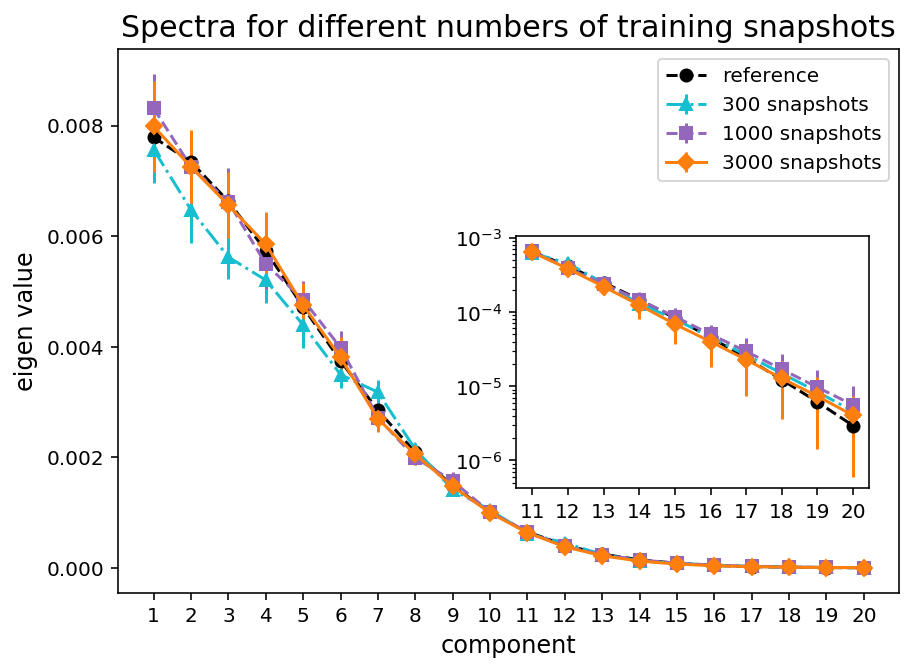}
	    \caption{}
	    \label{fig:1_PCA_2}
	\end{subfigure}	
	\caption{Forward problem: spectra of generated right-hand-side $f(x;\omega)$ processes. (a): Varying input noise dimension for a fixed number of snapshots at 1000. (b): Varying the number of training snapshots for a fixed dimension of input noise at 20. The inset plots are the zoom-ins of the eigenvalues of high frequency modes.  The means (markers) and two standard deviations (vertical lines on the markers) are calculated from the selected 33 generators. The reference curves are calculated from another $1\times 10^5$ independent $f(x;\omega)$ sample paths.}
	\label{fig:1_PCA}
\end{figure}

We also illustrate the spectra of the generated processes for $f(x;\omega)$ in Figure \ref{fig:1_PCA} to verify that the generated processes captured the covariance structure of $f(x;\omega)$. We can see that the spectra of our generated processes fit the reference solution well. With fixed number of training snapshots, as we increase the input noise dimension, the gap between our generated processes and the reference solutions narrows. With fixed input noise dimension, the gap narrows as we increase the number of training snapshots. These observations on the spectra are similar with those about the error of $u$.

We further check the correlation between the generated processes. The correlation coefficient $C(f_1(x;\omega),f_2(x;\omega))$ between two stochastic processes $f_1(x;\omega)$ and $f_2(x;\omega)$ ($x\in D$) is defined as 
\begin{equation}\label{eqn:corcoefficient}
C(f_1(x;\omega),f_2(x;\omega)) = \frac{1}{n}\sum_{i =1}^{n}\left|\frac{\mathrm{Cov}(f_1(x_i;\omega),f_2(x_i;\omega))}{\sqrt{\mathrm{Var}(f_1(x_i;\omega))\mathrm{Var}(f_2(x_i;\omega))}}\right|,
\end{equation}
where $x_i$ is the uniform grid points in $D$. Here, we set $n = 201$ for $C(k,f)$, while $n = 199$ for $C(k,u)$ to exclude the boundary points for $u$.
\begin{table}[H]
\centering
\begin{tabular}{| c || c | c | c | c |}
\hline
  & 2 dim & 4 dim & 20 dim & 40 dim \\ \hline
$C(k,f)$ & 3.4$\pm$2.1 \% & 2.3$\pm$1.6 \% & 2.5$\pm$2.3 \% & 2.7$\pm$1.8 \%  \\ \hline
$C(k,u)$   & 74.3$\pm$1.3 \%  & 73.6$\pm$1.0 \% & 73.8$\pm$ 0.9 \% & 73.2$\pm$0.8 \%  \\ \hline
\end{tabular}
\newline
\vspace*{0.4cm}
\newline
\begin{tabular}{| c || c | c | c |}
\hline
  & 300 snapshots & 3000 snapshots & reference  \\ \hline
$C(k,f)$ & 2.5$\pm$1.8 \% & 2.5$\pm$1.5 \% & 0 \\ \hline
$C(k,u)$   & 74.0$\pm$0.9 \% & 72.4$\pm$1.5 \% & 72.5 \%  \\ \hline 
\end{tabular}
\caption{Forward problem: correlation coefficient between the generated random processes. The mean and two standard deviations are calculated from the selected 33 generators for each case. The reference for $C(k,f)$ comes from the assumption that $k(x;\omega)$ and $f(x;\omega)$ are independent, while the reference for $C(k,u)$ is calculated from Equation (\ref{eqn:corcoefficient}) with $1\times 10^5$ independent samples of $k(x;\omega)$ and $u(x;\omega)$ paths.} 
\label{tab:case1_indep}
\end{table}

From Table~\ref{tab:case1_indep} we can see that our generated processes for $k(x;\omega)$ and $f(x;\omega)$ are weakly correlated, while the generated processes for $k(x;\omega)$ and $u(x;\omega)$ are strongly correlated. All the correlation coefficients are close to the reference values.

\subsubsection{Case 2: a (relatively) high dimensional problem}
\label{sec:high_forward}
In this case, we solve the stochastic differential equation where the correlation length of $f(x;\omega)$ is relatively small. We consider $k(x;\omega)$ and $f(x;\omega)$ as the following independent stochastic processes:
\begin{equation} \label{eqn:case2_samples}
\begin{aligned}
	 k(x) & = \exp[\frac{1}{5}\sin(\frac{3\pi}{2}(x +1)) + \hat{k}(x)] \\
	\hat{k}(x) & \sim \mathcal{GP}\left(0, \frac{4}{25}\exp(-(x-x')^2)\right) \\
	f(x) & \sim \mathcal{GP}\left(\frac{1}{2}, \frac{9}{400}\exp(\frac{-625(x-x')^2}{4})\right) \\
\end{aligned}
\end{equation}
Note that we need 30 dimensions to retain $99\%$ energy of $f(x;\omega)$. We put 13 $k$-sensors, 41 $f$-sensors, and 2 $u$-sensors on the boundary of our domain of interest $\mathcal{D}$. 
 In this case, we use $1\times10^4$ snapshots as training data while we set the batch size to be 1000 and fix the input noise dimension to be 20. Again, we shuffle the alignment of snapshots of $k(x;\omega)$ and $f(x;\omega)$ during the training to reflect their independence. 

\begin{figure}[H]
	\centering
	\begin{subfigure}[b]{0.3\textwidth}
	    \centering
	    \includegraphics[width=1.0\textwidth]{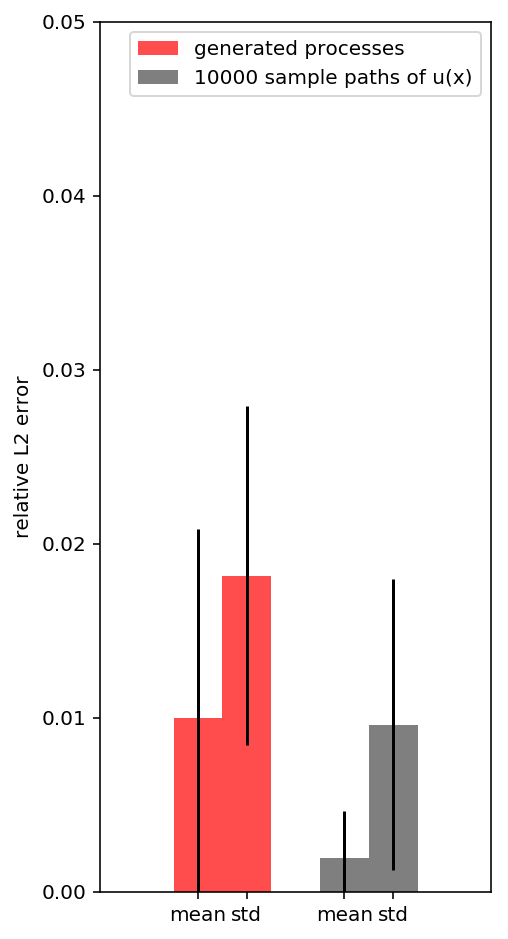}
	    \caption{}
	    \label{fig:2_error}
	\end{subfigure}%
	\begin{subfigure}[b]{0.7\textwidth}
	    \centering
	    \includegraphics[width=1.0\textwidth]{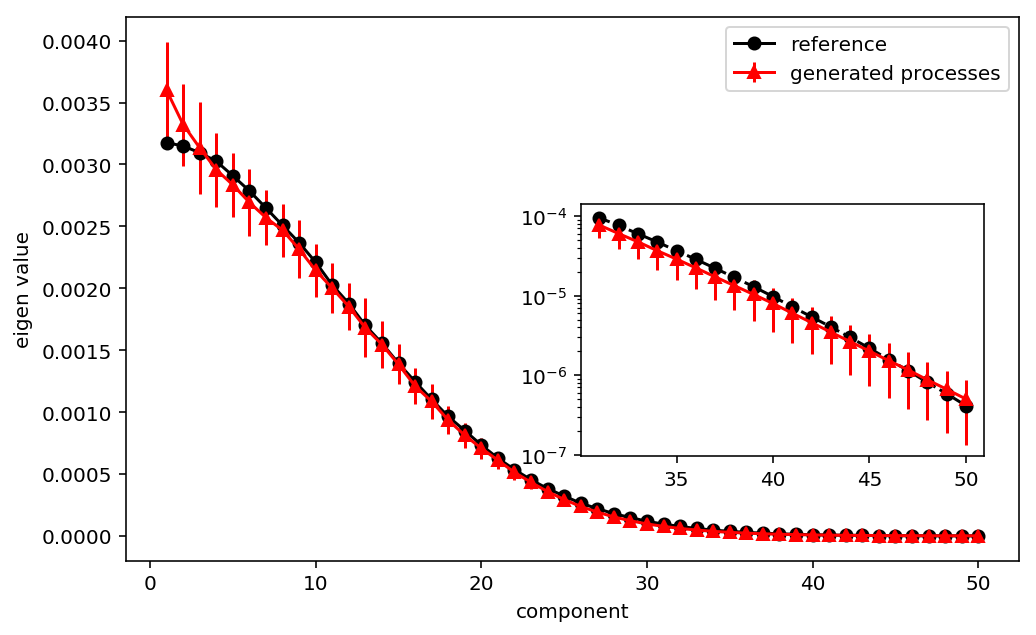}
	    \caption{}
	    \label{fig:2_PCA}
	\end{subfigure}
	\caption{Forward problem with stochastic right-hand-side process $f(x;\omega)$ of high dimensionality. (a) Relative errors of inferred mean and standard deviation of $u(x;\omega)$. Red bars and the associated black lines represent the mean and two standard deviations of relative errors from the selected 33 generators. The grey bars and the associated black lines represent the expectation and two standard deviations of relative errors calculated from $1\times10^4$ sample paths of $u(x;\omega)$. (b) The spectra of generated process versus the reference. The means (red markers) and two standard deviations (red vertical lines on the markers) are calculated from the selected 33 generators. The reference values are calculated from $1\times 10^5$ Monte Carlo sample paths of $f(x;\omega)$.}
	\label{fig:2_all}
\end{figure}

We run the code three times with different random seeds and stop the training after $2\times 10^5$ steps and then select 33 generators and generate $1\times 10^4$ sample paths from each generator in the same way as in Section \ref{sec:Toy} to calculate the error of $u(x;\omega)$ as well as the spectra of $f(x;\omega)$, as illustrated in Figure \ref{fig:2_all}. We can see that the spectra fit the reference values well. Although the error of our inferred mean is larger than that calculated from $1\times 10^4$ sample paths of $u(x;\omega)$, the error of standard deviation is comparable.

\subsection{Inverse and mixed problems}\label{sec:Inv}

In this section, we show that our method can manage a wide range of problems, from forward problems to inverse problems, and mixed problems in between. In particular, we solve the three types of problems governed by Equation (\ref{eqn:kufequation}), where $k(x;\omega)$ and $f(x;\omega)$ are independent processes as follows:
\begin{equation} \label{eqn:3_samples}
\begin{aligned}
	 k(x) & = \exp[\frac{1}{5}\sin(\frac{3\pi}{2}(x +1)) + \tilde{k}(x)], \\
	\tilde{k}(x) & \sim \mathcal{GP}\left(0, \frac{4}{25}\exp(-(x-x')^2)\right), \\
	f(x) & \sim \mathcal{GP}\left(\frac{1}{2}, \frac{9}{400}\exp(-(x-x')^2)\right). \\
\end{aligned}
\end{equation}
We consider the following four cases of sensor placement:
\begin{enumerate}[leftmargin=3\parindent]
    \item[Case 1] : 1 $k$-sensor, 13 $u$-sensors (including 2 on the boundary), 13 $f$-sensors.
    \item[Case 2] : 5 $k$-sensors, 9 $u$-sensors (including 2 on the boundary), 13 $f$-sensors.
    \item[Case 3] : 9 $k$-sensors, 5 $u$-sensors (including 2 on the boundary), 13 $f$-sensors.
    \item[Case 4] : 13 $k$-sensors, 2 $u$-sensors on the boundary of $u$, 13 $f$-sensors.
\end{enumerate}
Note that case 1 is an inverse problem, case 4 is a forward problem, while case 2 and 3 represent mixed problems. For each case, we use 1000 snapshots for training. Note that in cases 1-3, we \textit{cannot} shuffle the alignment of snapshots from $k(x;\omega)$ and $f(x;\omega)$ as in Section \ref{sec:Forward} since both $k(x;\omega)$ and $f(x;\omega)$ are correlated with $u(x;\omega)$. For consistency, in this section, we do not shuffle the alignment of snapshots for any of  the four cases. We set the batch size to be 1000 and the input noise dimension to be 20. For each case, we run the code three times with different random seeds. The training stops after $2\times 10^5$ steps, and we select 33 generators and generate $1\times 10^4$ sample paths from each generator in the same way as in Section \ref{sec:Toy} to calculate the mean and standard deviation of $k(x;\omega)$ and $u(x;\omega)$.

\begin{figure}[H]
	\centering
	\includegraphics[width=1.0\textwidth]{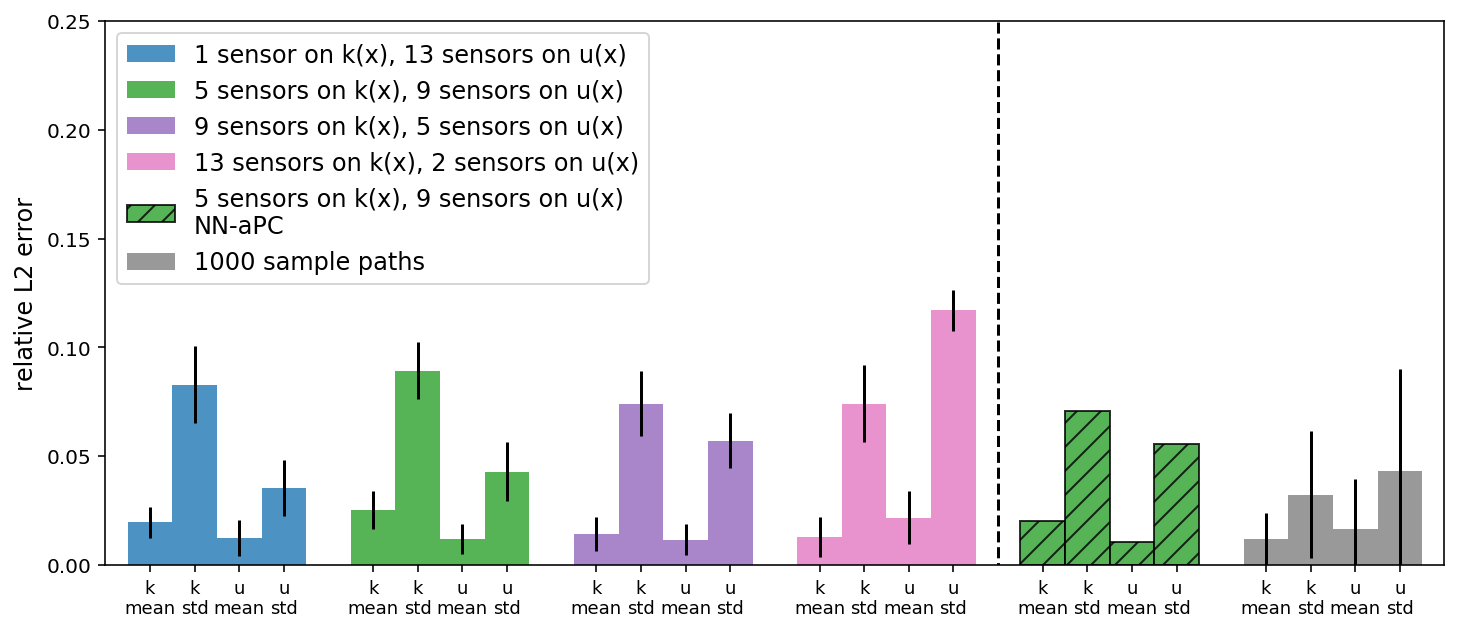}
	\caption{Relative errors of inferred mean and two standard deviations for both $k(x;\omega)$ and $u(x;\omega)$ in inverse problem (blue left most bars), mixed problems (green and purple bars) and forward problem (pink bars). The colored bars and the corresponding black lines represent the mean and two standard deviations of the relative errors calculated from the selected 33 generators for each case. Also shown on the right part of the panel are two reference cases, the green shaded bars corresponding to NN-aPC method \cite{zhang2018quantifying} for the case 2 on the left, and the grey bars and their associated black lines corresponding to the expectation and two standard deviations of relative error calculated from 1000 independent sample paths of $k(x;\omega)$ or $u(x;\omega)$.}
	\label{fig:3_error}
\end{figure}

	
In Figure \ref{fig:3_error} we compare the relative errors with reference solutions calculated from Monte Carlo sample paths as well as the method proposed  in~\cite{zhang2018quantifying}. We can see that our errors are in the same order of magnitude with errors from 1000 sample paths, showing the effectiveness of our method in solving all three types of problems. Also, for the case of 5 $k$-sensors and 9 $u$-sensors, our method achieves comparable accuracy with the method in \cite{zhang2018quantifying}.

\subsection{Multiple groups of training data}
Finally, we test our method for the case where we have multiple groups of snapshots as training data. In particular, we perform our test based on the case 4 in Section \ref{sec:Inv}. Apart from the sensors in that case, we put one additional $u$-sensor at position $x=0$, and collect another group of 1000 snapshots at this additional sensor. Hence, we have two groups of training data: 
\begin{enumerate}[leftmargin=3\parindent]
    \item[Group 1] : 13 $k$-sensors, 2 $u$-sensors on the boundary of $u$, 13 $f$-sensors. 1000 snapshots.
    \item[Group 2] : 1 $u$-sensor at $x=0$. 1000 snapshots.
\end{enumerate}
Again, we emphasize that the two groups of snapshots cannot be aligned. Note that a single snapshot in group 2 is almost useless since we cannot align it with another snapshot in group 1. However, the {\em ensemble} of the snapshots in group 2 actually can tell the {\em distribution} for $u(x;\omega)$ at $x=0$.

\begin{figure}[H]
	\centering
	\includegraphics[width=0.6\textwidth]{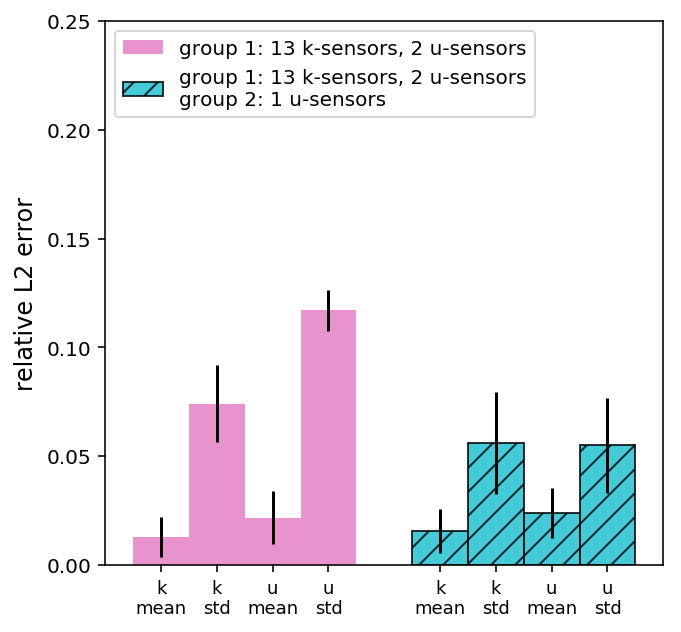}
	\caption{Relative errors of inferred mean and two standard deviations for both $k(x;\omega)$ and $u(x;\omega)$ using 1 group of data (pink) and 2 groups of data (light green). The colored bars and the corresponding black lines represent the mean and two standard deviations of the relative errors calculated from the selected 33 generators for each case.}
	\label{fig:4_error}
\end{figure}

To utilize the data in two groups, we apply Algorithm \ref{alg:SDE_multi} with the number of discriminators $M = 2$. In this case, we set the training batch size as 1000 while the input noise dimension as 20. We run the code three times with different random seeds. The training stops after $2\times 10^5$ steps, and we select 33 generators and generate $1\times 10^4$ sample paths from each generator in the same way as in Section \ref{sec:Toy} to calculate the mean and standard deviation of $k(x;\omega)$ and $u(x;\omega)$. In Figure \ref{fig:4_error}, we plot the errors of the inferred $k(x;\omega)$ and $u(x;\omega)$ in this case as well as errors from case 4 in Section \ref{sec:Inv}. Compared with the results obtained by only using one group of data, in this case, the error of the standard deviation of $u(x)$ decreases significantly, showing the capability of our method in {\em learning from the ensemble} of snapshots in group 2. To the best of our knowledge, our method is so far the only one that can manage this case.

\section{Summary and future work}\label{S:Conclusion}

We proposed physics-informed generative adversarial networks (PI-GANs) as a data-driven method for solving stochastic differential equations (SDEs) based on a limited number of scattered measurements. PI-GANs are composed of a discriminator, which is represented by a simple feed forward deep neural network (DNN) and of generators, which are a combination of feed forward DNNs and a neural network induced by the SDE.
We assumed that partial data are available in terms of different realizations of the stochastic process obtained simultaneously at different locations in the domain.  We trained the generators and discriminator iteratively with the loss functions employed in WGAN-GP \cite{gulrajani2017improved} so that the joint distribution of generated processes approximates the target stochastic processes. We also proposed a more general architecture with multiple discriminators to deal with cases, where data are collected in multiple groups, i.e., data collected independently from different sets of sensors.

We first tested WGAN-GP in approximating Gaussian processes for different correlation lengths. As shown in Figure \ref{fig:0_PCA}, we obtained good approximation of the generated stochastic processes to the target ones even for a mismatch between the input noise dimensionality and the effective dimensionality of the target stochastic processes. The approximations were improved by increasing the number of sensors and snapshots. We also compared WGAN-GP and vanilla GANs, and concluded that vanilla GANs are not suitable for approximating stochastic processes with deterministic boundary condition, as shown in Figure \ref{fig:0_compare}. We further studied the overfitting issue by monitoring the negative discriminator loss (Figure \ref{fig:4_loss}) and Wasserstein distance between empirical distributions (Figure \ref{fig:4_OT}). We found that overfitting occurs also in the generator in addition to the discriminator as previously reported. 

Subsequently, we considered the solution of elliptic SDEs requiring approximations of three stochastic processes, namely the solution $u(x;\omega)$, the forcing $f(x;\omega)$, and the diffusion coefficient $k(x;\omega)$. Without changing the framework, we were able to solve a wide range of problems, from forward to inverse problems, and in between, i.e., mixed problems where we have incomplete information for both the solution and the diffusion coefficient. As shown in Figure \ref{fig:1_MeanError_all} and Figure \ref{fig:3_error}, we obtained both the means and standard deviations of the stochastic solution and the diffusion coefficient in good agreement with benchmarks. In the case of the forward problem, we studied the influence of dimensionality of the input noise into the generators in Figure \ref{fig:1_MeanError_all}, and found that although small dimensionality can also work, high dimensionality for the input noise leads to better results. Moreover, we tested PI-GANs for a relatively high dimensional problem with $f(x;\omega)$ of stochastic dimension $30$. In Figure \ref{fig:2_all} we showed that the inferred mean and standard deviation of $u(x;\omega)$ as well as the spectra of $f(x;\omega)$ match the reference values well. Finally, we applied PI-GANs consisting of two discriminators to a stochastic problem with two groups of snapshots available for training, where the second group includes snapshots from only a single sensor for $u(x;\omega)$. We could see in Figure \ref{fig:4_error} that the error decreases compared with the error of the case where we only have the first group of data. This demonstrates the capability of PI-GANs to utilize information from multiple groups of data and learn from an \textit{ensemble} of snapshots, even when a single snapshot is useless.



We also point out some limitations of the current version of PI-GANs. Since the computational cost of training GANs is much higher than training a single feed forward neural network, the PI-GANs method has higher computational cost than the physics-informed neural networks for deterministic PDEs~\cite{MaziarParisGK17_1,MaziarParisGK17_2} and SDEs~\cite{zhang2018quantifying}. Also, in our numerical experiments, overfitting was detected in both discriminators and generators when training data are limited. Although overfitting of generators is less harmful than underfitting, we wish to address this issue systematically in future work. Moreover, in the current work we only take into consideration the uncertainty described in SDEs, but not from measurements, nor do we account for the uncertainty of the approximability of 
GANs as was done in ~\cite{zhang2018quantifying} for DNNs using the dropout method. In future work, we wish to endow PI-GANs with uncertainty quantification coming from diverse sources, and hence provide estimates of the total uncertainty of the unknown distribution. Finally, we comment on the computational cost due to high dimensionality of stochastic problems. In experiments not reported here, we estimated the computational cost of solving the forward SDE for 60 dimensions, i.e., twice the dimensionality in the case reported in Section \ref{sec:high_forward}. Specifically, we decreased the correlation length for $f(x;\omega)$ by half and doubled the sensors for $f(x;\omega)$, while keeping all the other settings the same as in Section \ref{sec:high_forward}. We found that the computational cost was approximately doubled, obtaining $1-3\%$ of relative error in the mean and standard deviation of the stochastic solution. This is consistent with other lower dimensional problems we presented in this paper, which suggests that the overall computational cost increases with a low-polynomial growth, hence PI-GANs can, in principle, tackle very high dimensional stochastic problems. We will systematically investigate the scalability of PI-GANs for more complex stochastic PDEs in very high dimensions in future work. Moreover, we shall aim to optimize PI-GANs in terms of the architecture, depth and width of the generators and discriminators as well as the other hyperparameters.


\section{Acknowledgement}
We would like to acknowledge support from the Army Research Office (ARO) W911NF-18-1-0301, and from the Department of Energy (DOE) DE-SC0019434 and DE-SC0019453.

\section*{References}
\bibliography{PI-GAN.bib}

\begin{thebibliography}{38}
\expandafter\ifx\csname natexlab\endcsname\relax\def\natexlab#1{#1}\fi
\providecommand{\bibinfo}[2]{#2}
\ifx\xfnm\relax \def\xfnm[#1]{\unskip,\space#1}\fi
\bibitem[{Berthelot et~al.(2017)Berthelot, Schumm, and
  Metz}]{berthelot2017began}
\bibinfo{author}{D.~Berthelot}, \bibinfo{author}{T.~Schumm},
  \bibinfo{author}{L.~Metz},
\newblock \bibinfo{title}{{BEGAN}: {B}oundary equilibrium generative
  adversarial networks},
\newblock \bibinfo{journal}{arXiv preprint}  (\bibinfo{year}{2017})
  \bibinfo{pages}{arXiv:1703.10717}.
\bibitem[{Karras et~al.(2017)Karras, Aila, Laine, and
  Lehtinen}]{karras2017progressive}
\bibinfo{author}{T.~Karras}, \bibinfo{author}{T.~Aila},
  \bibinfo{author}{S.~Laine}, \bibinfo{author}{J.~Lehtinen},
\newblock \bibinfo{title}{Progressive growing of {GANs} for improved quality,
  stability, and variation},
\newblock \bibinfo{journal}{arXiv preprint}  (\bibinfo{year}{2017})
  \bibinfo{pages}{arXiv:1710.10196}.
\bibitem[{Ledig et~al.(????)Ledig, Theis, Husz{\'a}r, Caballero, Cunningham,
  Acosta, Aitken, Tejani, Totz, Wang et~al.}]{ledig2017photo}
\bibinfo{author}{C.~Ledig}, \bibinfo{author}{L.~Theis},
  \bibinfo{author}{F.~Husz{\'a}r}, \bibinfo{author}{J.~Caballero},
  \bibinfo{author}{A.~Cunningham}, \bibinfo{author}{A.~Acosta},
  \bibinfo{author}{A.~P. Aitken}, \bibinfo{author}{A.~Tejani},
  \bibinfo{author}{J.~Totz}, \bibinfo{author}{Z.~Wang}, et~al.,
\newblock \bibinfo{title}{Photo-realistic single image super-resolution using a
  generative adversarial network},
\newblock in: \bibinfo{booktitle}{CVPR (2017)}, p.~\bibinfo{pages}{4}.
\bibitem[{Zhu et~al.(2017)Zhu, Park, Isola, and Efros}]{CycleGAN2017}
\bibinfo{author}{J.-Y. Zhu}, \bibinfo{author}{T.~Park},
  \bibinfo{author}{P.~Isola}, \bibinfo{author}{A.~A. Efros},
\newblock \bibinfo{title}{Unpaired image-to-image translation using
  cycle-consistent adversarial networks},
\newblock \bibinfo{journal}{arXiv preprint}  (\bibinfo{year}{2017})
  \bibinfo{pages}{arXiv:1703.10593}.
\bibitem[{Yu et~al.(????)Yu, Zhang, Wang, and Yu}]{yu2017seqgan}
\bibinfo{author}{L.~Yu}, \bibinfo{author}{W.~Zhang}, \bibinfo{author}{J.~Wang},
  \bibinfo{author}{Y.~Yu},
\newblock \bibinfo{title}{Seq{GAN}: Sequence generative adversarial nets with
  policy gradient},
\newblock in: \bibinfo{booktitle}{AAAI (2017)}, pp.
  \bibinfo{pages}{2852--2858}.
\bibitem[{Zhang et~al.(2017)Zhang, Gan, Fan, Chen, Henao, Shen, and
  Carin}]{zhang2017adversarial}
\bibinfo{author}{Y.~Zhang}, \bibinfo{author}{Z.~Gan}, \bibinfo{author}{K.~Fan},
  \bibinfo{author}{Z.~Chen}, \bibinfo{author}{R.~Henao},
  \bibinfo{author}{D.~Shen}, \bibinfo{author}{L.~Carin},
\newblock \bibinfo{title}{Adversarial feature matching for text generation},
\newblock \bibinfo{journal}{arXiv preprint}  (\bibinfo{year}{2017})
  \bibinfo{pages}{arXiv:1706.03850}.
\bibitem[{Fedus et~al.(2018)Fedus, Goodfellow, and Dai}]{fedus2018maskgan}
\bibinfo{author}{W.~Fedus}, \bibinfo{author}{I.~Goodfellow},
  \bibinfo{author}{A.~M. Dai},
\newblock \bibinfo{title}{Mask{GAN}: Better text generation via filling in the
  \_\_\_\_\_\_},
\newblock \bibinfo{journal}{arXiv preprint}  (\bibinfo{year}{2018})
  \bibinfo{pages}{arXiv:1801.07736}.
\bibitem[{Liang et~al.(2017)Liang, Hu, Zhang, Gan, and
  Xing}]{liang2017recurrent}
\bibinfo{author}{X.~Liang}, \bibinfo{author}{Z.~Hu},
  \bibinfo{author}{H.~Zhang}, \bibinfo{author}{C.~Gan}, \bibinfo{author}{E.~P.
  Xing},
\newblock \bibinfo{title}{Recurrent topic-transition {GAN} for visual paragraph
  generation},
\newblock \bibinfo{journal}{arXiv preprint}  (\bibinfo{year}{2017})
  \bibinfo{pages}{arXiv:1703.07022}.
\bibitem[{Mogren(2016)}]{mogren2016c}
\bibinfo{author}{O.~Mogren},
\newblock \bibinfo{title}{{C-RNN-GAN}: Continuous recurrent neural networks
  with adversarial training},
\newblock \bibinfo{journal}{arXiv preprint}  (\bibinfo{year}{2016})
  \bibinfo{pages}{arXiv:1611.09904}.
\bibitem[{Yang et~al.(2017)Yang, Chou, and Yang}]{yang2017midinet}
\bibinfo{author}{L.-C. Yang}, \bibinfo{author}{S.-Y. Chou},
  \bibinfo{author}{Y.-H. Yang},
\newblock \bibinfo{title}{Midinet: A convolutional generative adversarial
  network for symbolic-domain music generation},
\newblock \bibinfo{journal}{arXiv preprint}  (\bibinfo{year}{2017})
  \bibinfo{pages}{arXiv:1703.10847}.
\bibitem[{Guimaraes et~al.(2017)Guimaraes, Sanchez-Lengeling, Outeiral, Farias,
  and Aspuru-Guzik}]{guimaraes2017objective}
\bibinfo{author}{G.~L. Guimaraes}, \bibinfo{author}{B.~Sanchez-Lengeling},
  \bibinfo{author}{C.~Outeiral}, \bibinfo{author}{P.~L.~C. Farias},
  \bibinfo{author}{A.~Aspuru-Guzik},
\newblock \bibinfo{title}{Objective-reinforced generative adversarial networks
  {(ORGAN)} for sequence generation models},
\newblock \bibinfo{journal}{arXiv preprint}  (\bibinfo{year}{2017})
  \bibinfo{pages}{arXiv:1705.10843}.
\bibitem[{{Raissi} et~al.(2017{\natexlab{a}}){Raissi}, {Perdikaris}, and
  {Karniadakis}}]{MaziarParisGK17_1}
\bibinfo{author}{M.~{Raissi}}, \bibinfo{author}{P.~{Perdikaris}},
  \bibinfo{author}{G.~E. {Karniadakis}},
\newblock \bibinfo{title}{{Physics informed deep learning (part I): Data-driven
  solutions of nonlinear partial differential equations}},
\newblock \bibinfo{journal}{arXiv preprint}
  (\bibinfo{year}{2017}{\natexlab{a}}) \bibinfo{pages}{arXiv:1711.10561}.
\bibitem[{{Raissi} et~al.(2017{\natexlab{b}}){Raissi}, {Perdikaris}, and
  {Karniadakis}}]{MaziarParisGK17_2}
\bibinfo{author}{M.~{Raissi}}, \bibinfo{author}{P.~{Perdikaris}},
  \bibinfo{author}{G.~E. {Karniadakis}},
\newblock \bibinfo{title}{{Physics informed deep learning (part II):
  Data-driven discovery of nonlinear partial differential equations}},
\newblock \bibinfo{journal}{arXiv preprint}
  (\bibinfo{year}{2017}{\natexlab{b}}) \bibinfo{pages}{arXiv:1711.10566}.
\bibitem[{Schmidt and Lipson(2009)}]{schmidt2009distilling}
\bibinfo{author}{M.~Schmidt}, \bibinfo{author}{H.~Lipson},
\newblock \bibinfo{title}{Distilling free-form natural laws from experimental
  data},
\newblock \bibinfo{journal}{Science} \bibinfo{volume}{324}
  (\bibinfo{year}{2009}) \bibinfo{pages}{81--85}.
\bibitem[{Brunton et~al.(2016)Brunton, Proctor, and
  Kutz}]{brunton2016discovering}
\bibinfo{author}{S.~L. Brunton}, \bibinfo{author}{J.~L. Proctor},
  \bibinfo{author}{J.~N. Kutz},
\newblock \bibinfo{title}{Discovering governing equations from data by sparse
  identification of nonlinear dynamical systems},
\newblock \bibinfo{journal}{Proceedings of the National Academy of Sciences}
  (\bibinfo{year}{2016}) \bibinfo{pages}{201517384}.
\bibitem[{Raissi and Karniadakis(2018)}]{MaziarGK18JCP}
\bibinfo{author}{M.~Raissi}, \bibinfo{author}{G.~E. Karniadakis},
\newblock \bibinfo{title}{{Hidden physics models: Machine learning of nonlinear
  partial differential equations}},
\newblock \bibinfo{journal}{Journal of Computational Physics}
  \bibinfo{volume}{357} (\bibinfo{year}{2018}) \bibinfo{pages}{125--141}.
\bibitem[{Graepel(2003)}]{graepel2003}
\bibinfo{author}{T.~Graepel},
\newblock \bibinfo{title}{{Solving noisy linear operator equations by
  {Gaussian} processes: Application to ordinary and partial differential
  equations}},
\newblock in: \bibinfo{booktitle}{International Conference on Machine Learning
  (2003)}, pp. \bibinfo{pages}{234--241}.
\bibitem[{S{\"a}rkk{\"a}(2011)}]{sarkka2011}
\bibinfo{author}{S.~S{\"a}rkk{\"a}},
\newblock \bibinfo{title}{Linear operators and stochastic partial differential
  equations in {Gaussian} process regression},
\newblock in: \bibinfo{booktitle}{International Conference on Artificial Neural
  Networks (2011)}, \bibinfo{organization}{Springer}, pp.
  \bibinfo{pages}{151--158}.
\bibitem[{Bilionis(2016)}]{bilionis2016}
\bibinfo{author}{I.~Bilionis},
\newblock \bibinfo{title}{{Probabilistic solvers for partial differential
  equations}},
\newblock \bibinfo{journal}{arXiv preprint}  (\bibinfo{year}{2016})
  \bibinfo{pages}{arXiv:1607.03526}.
\bibitem[{Raissi et~al.(2018)Raissi, Perdikaris, and
  Karniadakis}]{Raissi_nonlinear}
\bibinfo{author}{M.~Raissi}, \bibinfo{author}{P.~Perdikaris},
  \bibinfo{author}{G.~E. Karniadakis},
\newblock \bibinfo{title}{Numerical {Gaussian} processes for time-dependent and
  nonlinear partial differential equations},
\newblock \bibinfo{journal}{SIAM Journal on Scientific Computing}
  \bibinfo{volume}{40} (\bibinfo{year}{2018}) \bibinfo{pages}{A172--A198}.
\bibitem[{Pang et~al.(2018)Pang, Yang, and Karniadakis}]{Pang2018}
\bibinfo{author}{G.~Pang}, \bibinfo{author}{L.~Yang}, \bibinfo{author}{G.~E.
  Karniadakis},
\newblock \bibinfo{title}{{Neural-net-induced Gaussian process regression for
  function approximation and PDE solution}},
\newblock \bibinfo{journal}{arXiv preprint}  (\bibinfo{year}{2018})
  \bibinfo{pages}{arXiv:1806.11187}.
\bibitem[{Yang et~al.(2018)Yang, Tartakovsky, and Tartakovsky}]{xiuyang_GP}
\bibinfo{author}{X.~Yang}, \bibinfo{author}{G.~Tartakovsky},
  \bibinfo{author}{A.~Tartakovsky},
\newblock \bibinfo{title}{{Physics-informed kriging: A physics-informed
  Gaussian process regression method for data-model convergence}},
\newblock \bibinfo{journal}{arXiv preprint}  (\bibinfo{year}{2018})
  \bibinfo{pages}{arXiv:1809.03461}.
\bibitem[{Lagaris et~al.(1998)Lagaris, Likas, and Fotiadis}]{Lagaris1997}
\bibinfo{author}{I.~E. Lagaris}, \bibinfo{author}{A.~C. Likas},
  \bibinfo{author}{D.~I. Fotiadis},
\newblock \bibinfo{title}{Artificial neural networks for solving ordinary and
  partial differential equations},
\newblock \bibinfo{journal}{IEEE Transactions on Neural Networks}
  \bibinfo{volume}{9} (\bibinfo{year}{1998}) \bibinfo{pages}{987--1000}.
\bibitem[{Lagaris et~al.(2000)Lagaris, Likas, and Papageorgiou}]{Lagaris2000}
\bibinfo{author}{I.~E. Lagaris}, \bibinfo{author}{A.~C. Likas},
  \bibinfo{author}{D.~G. Papageorgiou},
\newblock \bibinfo{title}{Neural-network methods for boundary value problems
  with irregular boundaries},
\newblock \bibinfo{journal}{IEEE Transactions on Neural Networks}
  \bibinfo{volume}{11} (\bibinfo{year}{2000}) \bibinfo{pages}{1041--1049}.
\bibitem[{Khoo et~al.(2017)Khoo, Lu, and Ying}]{Yinglexing17}
\bibinfo{author}{Y.~Khoo}, \bibinfo{author}{J.~Lu}, \bibinfo{author}{L.~Ying},
\newblock \bibinfo{title}{Solving parametric {PDE} problems with artificial
  neural networks},
\newblock \bibinfo{journal}{arXiv preprint}  (\bibinfo{year}{2017})
  \bibinfo{pages}{arXiv:1707.03351}.
\bibitem[{Nabian and Meidani(2018)}]{Hadi18}
\bibinfo{author}{M.~A. Nabian}, \bibinfo{author}{H.~Meidani},
\newblock \bibinfo{title}{A deep neural network surrogate for high-dimensional
  random partial differential equations},
\newblock \bibinfo{journal}{arXiv preprint}  (\bibinfo{year}{2018})
  \bibinfo{pages}{arXiv:1806.02957}.
\bibitem[{Stuart(2010)}]{Stuart10}
\bibinfo{author}{A.~M. Stuart},
\newblock \bibinfo{title}{{Inverse problems: A Bayesian perspective}},
\newblock \bibinfo{journal}{Acta Numerica} \bibinfo{volume}{19}
  (\bibinfo{year}{2010}) \bibinfo{pages}{451--559}.
\bibitem[{Raissi et~al.(2017)Raissi, Perdikaris, and
  Karniadakis}]{raissi_jcp_2017}
\bibinfo{author}{M.~Raissi}, \bibinfo{author}{P.~Perdikaris},
  \bibinfo{author}{G.~E. Karniadakis},
\newblock \bibinfo{title}{Machine learning of linear differential equations
  using {Gaussian} processes},
\newblock \bibinfo{journal}{Journal of Computational Physics}
  \bibinfo{volume}{348} (\bibinfo{year}{2017}) \bibinfo{pages}{683--693}.
\bibitem[{Zhu and Zabaras(2018)}]{zhu_zabaras18}
\bibinfo{author}{Y.~Zhu}, \bibinfo{author}{N.~Zabaras},
\newblock \bibinfo{title}{{Bayesian deep convolutional encoder-decoder networks
  for surrogate modeling and uncertainty quantification}},
\newblock \bibinfo{journal}{Journal of Computational Physics}
  \bibinfo{volume}{366} (\bibinfo{year}{2018}) \bibinfo{pages}{415--447}.
\bibitem[{{E} et~al.(2017){E}, {Han}, and {Jentzen}}]{Weinan-arxiv}
\bibinfo{author}{W.~{E}}, \bibinfo{author}{J.~{Han}},
  \bibinfo{author}{A.~{Jentzen}},
\newblock \bibinfo{title}{{Deep learning-based numerical methods for
  high-dimensional parabolic partial differential equations and backward
  stochastic differential equations}},
\newblock \bibinfo{journal}{arXiv preprint}  (\bibinfo{year}{2017})
  \bibinfo{pages}{arXiv:1706.04702}.
\bibitem[{{Raissi}(2018)}]{Maziar-arxiv}
\bibinfo{author}{M.~{Raissi}},
\newblock \bibinfo{title}{{Forward-backward stochastic neural networks: Deep
  learning of high-dimensional partial differential equations}},
\newblock \bibinfo{journal}{arXiv preprint}  (\bibinfo{year}{2018})
  \bibinfo{pages}{arXiv:1804.07010}.
\bibitem[{Zhang et~al.(2018)Zhang, Lu, Guo, and
  Karniadakis}]{zhang2018quantifying}
\bibinfo{author}{D.~Zhang}, \bibinfo{author}{L.~Lu}, \bibinfo{author}{L.~Guo},
  \bibinfo{author}{G.~E. Karniadakis},
\newblock \bibinfo{title}{Quantifying total uncertainty in physics-informed
  neural networks for solving forward and inverse stochastic problems},
\newblock \bibinfo{journal}{arXiv preprint}  (\bibinfo{year}{2018})
  \bibinfo{pages}{arXiv:1809.08327}.
\bibitem[{Goodfellow et~al.(2014)Goodfellow, Pouget-Abadie, Mirza, Xu,
  Warde-Farley, Ozair, Courville, and Bengio}]{goodfellow2014generative}
\bibinfo{author}{I.~Goodfellow}, \bibinfo{author}{J.~Pouget-Abadie},
  \bibinfo{author}{M.~Mirza}, \bibinfo{author}{B.~Xu},
  \bibinfo{author}{D.~Warde-Farley}, \bibinfo{author}{S.~Ozair},
  \bibinfo{author}{A.~Courville}, \bibinfo{author}{Y.~Bengio},
\newblock \bibinfo{title}{Generative adversarial nets},
\newblock in: \bibinfo{booktitle}{Advances in neural information processing
  systems (2014)}, pp. \bibinfo{pages}{2672--2680}.
\bibitem[{Arjovsky et~al.(2017)Arjovsky, Chintala, and
  Bottou}]{arjovsky2017wasserstein}
\bibinfo{author}{M.~Arjovsky}, \bibinfo{author}{S.~Chintala},
  \bibinfo{author}{L.~Bottou},
\newblock \bibinfo{title}{Wasserstein {GAN}},
\newblock \bibinfo{journal}{arXiv preprint}  (\bibinfo{year}{2017})
  \bibinfo{pages}{arXiv:1701.07875}.
\bibitem[{Gulrajani et~al.(2017)Gulrajani, Ahmed, Arjovsky, Dumoulin, and
  Courville}]{gulrajani2017improved}
\bibinfo{author}{I.~Gulrajani}, \bibinfo{author}{F.~Ahmed},
  \bibinfo{author}{M.~Arjovsky}, \bibinfo{author}{V.~Dumoulin},
  \bibinfo{author}{A.~C. Courville},
\newblock \bibinfo{title}{Improved training of {Wasserstein GANs}},
\newblock in: \bibinfo{booktitle}{Advances in Neural Information Processing
  Systems (2017)}, pp. \bibinfo{pages}{5767--5777}.
\bibitem[{Kingma and Ba(2014)}]{kingma2014adam}
\bibinfo{author}{D.~P. Kingma}, \bibinfo{author}{J.~Ba},
\newblock \bibinfo{title}{Adam: A method for stochastic optimization},
\newblock \bibinfo{journal}{arXiv preprint}  (\bibinfo{year}{2014})
  \bibinfo{pages}{arXiv:1412.6980}.
\bibitem[{Baydin et~al.(2017)Baydin, Pearlmutter, Radul, and
  Siskind}]{baydin2017automatic}
\bibinfo{author}{A.~G. Baydin}, \bibinfo{author}{B.~A. Pearlmutter},
  \bibinfo{author}{A.~A. Radul}, \bibinfo{author}{J.~M. Siskind},
\newblock \bibinfo{title}{Automatic differentiation in machine learning: a
  survey},
\newblock \bibinfo{journal}{Journal of Machine Learning Research}
  \bibinfo{volume}{18} (\bibinfo{year}{2017}) \bibinfo{pages}{1--153}.
\bibitem[{Flamary and Courty(2017)}]{flamary2017pot}
\bibinfo{author}{R.~Flamary}, \bibinfo{author}{N.~Courty},
  \bibinfo{title}{{POT} {Python} optimal transport library},
  \bibinfo{year}{(2017)}.

\end{thebibliography}

\end{document}